%% file: main.tex
\documentclass[10pt,twocolumn,letterpaper]{article}
\usepackage[pagenumbers]{cvpr}              %
\usepackage[accsupp]{axessibility}

\PassOptionsToPackage{hypcap=false}{caption}

\input{preamble}
\definecolor{cvprblue}{rgb}{0.21,0.49,0.74}
\definecolor{strongred}{rgb}{1.0, 0.65, 0.65} 
\definecolor{lightred}{rgb}{1.0, 0.82, 0.82} 
\definecolor{strongblue}{rgb}{0.7, 0.85, 1.0}
\definecolor{lightblue}{rgb}{0.8, 0.9, 1.0}
\usepackage{booktabs}
\usepackage{multirow}
\usepackage{colortbl}
\usepackage{xcolor}
\usepackage{fontawesome}
\usepackage[pagebackref,breaklinks,colorlinks,allcolors=cvprblue]{hyperref}
\def\paperID{16} %
\def\confName{CVPR Workshop}
\def\confYear{2025}
\title{SVAD: From Single Image to 3D Avatar via Synthetic Data Generation with Video Diffusion and Data Augmentation}
\author{Yonwoo Choi\\
SECERN AI\\
{\tt\small yonwoo.choi@secern.ai} \\
{\tt\small \href{https://yc4ny.github.io/SVAD/}{\color{magenta}{https://yc4ny.github.io/SVAD/}}}
}
\begin{document}
\twocolumn[
{
\renewcommand\twocolumn[1][]{#1}%
\maketitle
\vspace{-2.5em}  %
\centering  %
\includegraphics[width=1\linewidth]{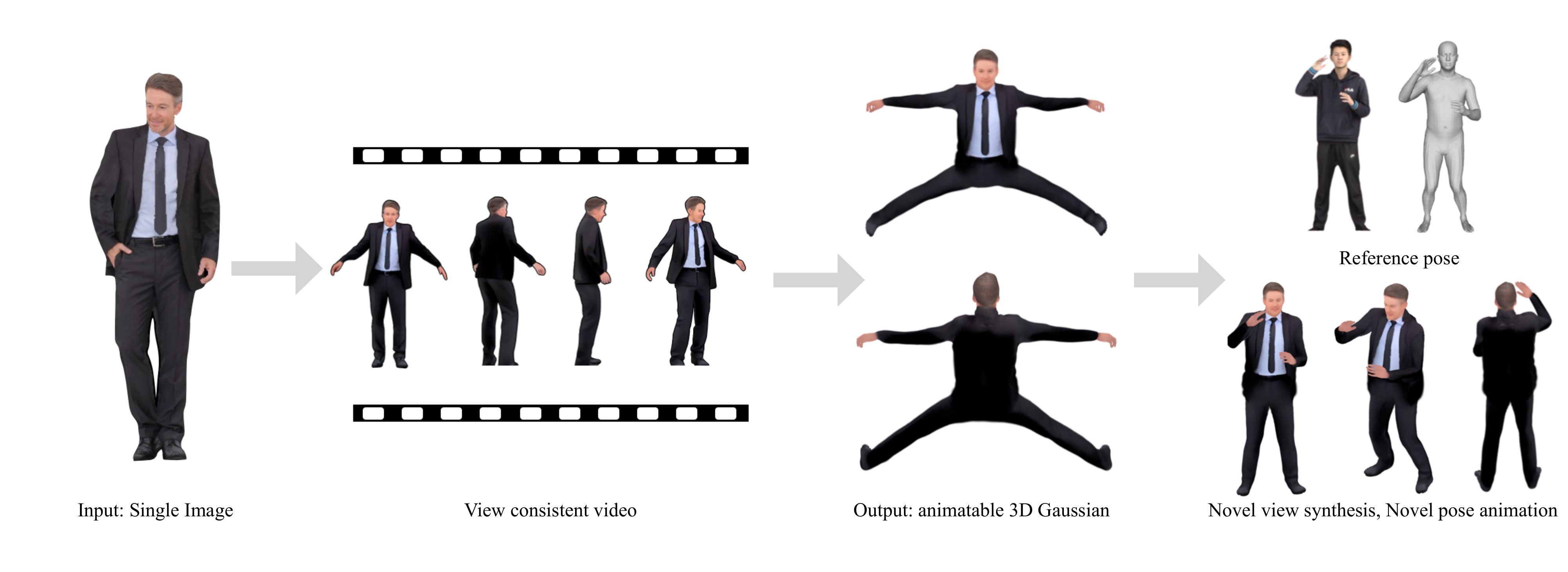}
\vspace{-2em}  %
\captionsetup{hypcap=false}  %
\captionof{figure}{\textbf{SVAD.} Our method creates high-fidelity 3D avatars from a single image through synthetic data generation. We leverage video diffusion to generate pose-conditioned animations, enhance them with identity preservation and image restoration modules, then train a 3D Gaussian Splatting avatar. The resulting avatars maintain consistent identity across novel poses and viewpoints while enabling real-time rendering, outperforming state-of-the-art approaches.}
\label{fig:teaser}
\vspace{1em}
}
]
\maketitle
\input{sec/0_abstract}    
\input{sec/1_intro}
\input{sec/2_related_work}
\input{sec/3_method}

\input{sec/4_experiments}
\input{sec/5_conclusion}
\clearpage
{
    \small
    \bibliographystyle{ieeenat_fullname}
    \bibliography{main}
}
\setcounter{section}{0} 
\renewcommand{\thesection}{\Alph{section}} 
\input{sec/X_suppl}
\end{document}

%% file: sec/0_abstract.tex
\begin{abstract}
Creating high-quality animatable 3D human avatars from a single image remains a significant challenge in computer vision due to the inherent difficulty of reconstructing complete 3D information from a single viewpoint. Current approaches face a clear limitation: 3D Gaussian Splatting (3DGS) methods produce high-quality results but require multiple views or video sequences, while video diffusion models can generate animations from single images but struggle with consistency and identity preservation. We present SVAD, a novel approach that addresses these limitations by leveraging complementary strengths of existing techniques. Our method generates synthetic training data through video diffusion, enhances it with identity preservation and image restoration modules, and utilizes this refined data to train 3DGS avatars. Comprehensive evaluations demonstrate that SVAD outperforms state-of-the-art (SOTA) single-image methods in maintaining identity consistency and fine details across novel poses and viewpoints, while enabling real-time rendering capabilities. Through our data augmentation pipeline, we overcome the dependency on dense monocular or multi-view training data typically required by traditional 3DGS approaches. Extensive quantitative, qualitative comparisons show our method achieves superior performance across multiple metrics against baseline models. By effectively combining the generative power of diffusion models with both the high-quality results and rendering efficiency of 3DGS, our work establishes a new approach for high-fidelity avatar generation from a single image input.
\end{abstract}

%% file: sec/1_intro.tex
\section{Introduction} \label{sec:introduction}
The ability to generate animatable 3D human avatars from minimal input data, such as a single-image, has significant potential across a range of applications. Traditional methods, particularly those based on 3DGS, have demonstrated considerable success in producing high-quality avatars~\cite{hu2024gaussianavatar, moon2024expressive, shao2024splattingavatar, qian20243dgs, moreau2024human, svitov2024haha, shen2023x, zeng2023avatarbooth, chen2024meshavatar, zheng2024physavatar}. These methods rely on dense input data, typically monocular or multi-view video~\cite{moon2024expressive, hu2024gaussianavatar, qian20243dgs, shao2024splattingavatar, svitov2024haha, chen2024meshavatar, zheng2024physavatar}, to achieve high fidelity across varied viewpoints and poses. This reliance on extensive video input complicates deployment in single-image scenarios, where ensuring viewpoint consistency and adaptability to novel poses becomes a key challenge.

Recent advancements in video diffusion models offer a potential solution by enabling animation generation from a single static image~\cite{zhu2024champ, hu2024animate, xu2024magicanimate, wang2024disco, kim2025target}. These models use certain conditions in diffusion processes to create video sequences, demonstrating the powerful generative capabilities of diffusion for single-image-driven animation. However, diffusion models often struggle to maintain temporal coherence, leading to inconsistent features and identity drift across frames~\cite{singer2022make,ho2022video,blattmann2023align,ding2023diffusionrig} . Additionally, their iterative denoising process for each frame introduces significant computational overhead, limiting their feasibility for real-time or interactive applications where rapid rendering across novel views is essential.

To overcome these challenges, we propose SVAD, a novel synthetic data generation and avatar creation pipeline that synergizes the generative flexibility of diffusion models with the efficient rendering capabilities of 3DGS avatars. Our approach leverages video diffusion model~\cite{musepose} to generate diverse pose-conditioned synthetic training data from a single-image. This synthetic data is refined through an identity-preservation module and an image restoration module to ensure that perceptual identity consistency and structural fidelity are preserved across diverse poses and temporal sequences. The resulting high-quality synthetic dataset is then used to train a 3DGS avatar model~\cite{moon2024expressive}, which benefits from the rapid rendering capabilities inherent to 3DGS. By combining the generative strengths of diffusion for synthetic data creation with the efficiency of 3DGS for rendering, SVAD achieves consistent, high-quality 3D avatar animations from single-image input.

In summary, our main contributions are:
\begin{itemize}
    \item We introduce a novel pipeline that generates high-quality synthetic training data from a single-image to create detailed, animatable 3D human avatars.
    
    \item We develop a comprehensive data augmentation approach that combines identity preservation and image restoration to ensure consistent identity and fine details across diverse poses.
    
    \item We demonstrate through extensive experiments that our synthetic data-driven approach significantly outperforms SOTA single-image avatar generation methods in identity preservation and novel pose adaptation while maintaining efficient real-time rendering.
\end{itemize}

%% file: sec/2_related_work.tex
\section{Related Work}
\label{sec:related_work}

\noindent\textbf{Diffusion Model for Human Image Animation}
The use of diffusion models has led to significant advancements in human image animation, enabling the generation of realistic and temporally consistent animations from static images~\cite{zhao2022thin,albahar2023single,cao2024dreamavatar,chan2019everybody,fu2022stylegan, hu2023sherf,jiang2023humangen, prokudin2021smplpix, ren2020deep,sarkar2020neural, siarohin2019first, siarohin2021motion, yoon2021pose, yu2023bidirectionally, zhang2022exploring}. Early methods, such as PIDM~\cite{bhunia2023person} and DreamPose~\cite{karras2023dreampose}, focused on improving texture fidelity by employing texture diffusion modules to align texture patterns between reference and target images. These methods, while enhancing detail preservation, still face challenges in maintaining temporal stability across frames.

Recent works, including DisCo~\cite{wang2024disco} and Animate Anyone~\cite{hu2024animate}, have extended diffusion models to improve temporal consistency and fine-grained control in human animation tasks. DisCo leverages dual ControlNets~\cite{zhang2023adding} to separately control pose and background elements, providing more robust conditioning for complex motion sequences. Similarly, Animate Anyone integrates a ReferenceNet with temporal attention layers to ensure appearance consistency and smooth transitions across frames, thereby addressing flickering issues commonly observed in earlier models. 

\noindent\textbf{Dynamic 3D Gaussian based Avatars}
The concept of Gaussian splatting for 3D avatars has emerged recently as an innovative approach to explicit scene representation~\cite{kerbl20233d}. This technique models a scene as a collection of 3D Gaussian elements, each containing photometric and geometric properties. During rendering, these Gaussian splats are projected onto the image plane, creating the final rendered output. The efficiency of 3DGS has been demonstrated in both static~\cite{huang2024error,jiang2024gaussianshader, lee2024deblurring} and dynamic~\cite{duan20244d, jones2021belonging, kratimenos2025dynmf, luiten2023dynamic,lee2024guess} scenes, making it a versatile tool for various applications.
Recent advancements~\cite{chen2024monogaussianavatar, dhamo2025headgas, hu2024gauhuman, jena2023splatarmor, lei2024gart, qian2024gaussianavatars,qian20243dgs, wang2023gaussianhead, zielonka2023drivable, cha2024pegasus,cha2024perse} have explored the use of 3DGS to create photorealistic human avatars across different scenarios. These methods commonly rely on multi-view data~\cite{li2024animatable, pang2024ash, zheng2024gps} or monocular video~\cite{hu2024gaussianavatar, jena2023splatarmor,moon2024expressive,lei2024gart,qian20243dgs} as input to achieve high-quality, consistent results. The advantage of 3DGS lies in its ability to produce temporally stable animated avatars with superior quantitative metrics.

%% file: sec/3_method.tex
\section{Method}
\label{sec:method}

\begin{figure*}[ht]
\centering
\includegraphics[width=1.0\linewidth]{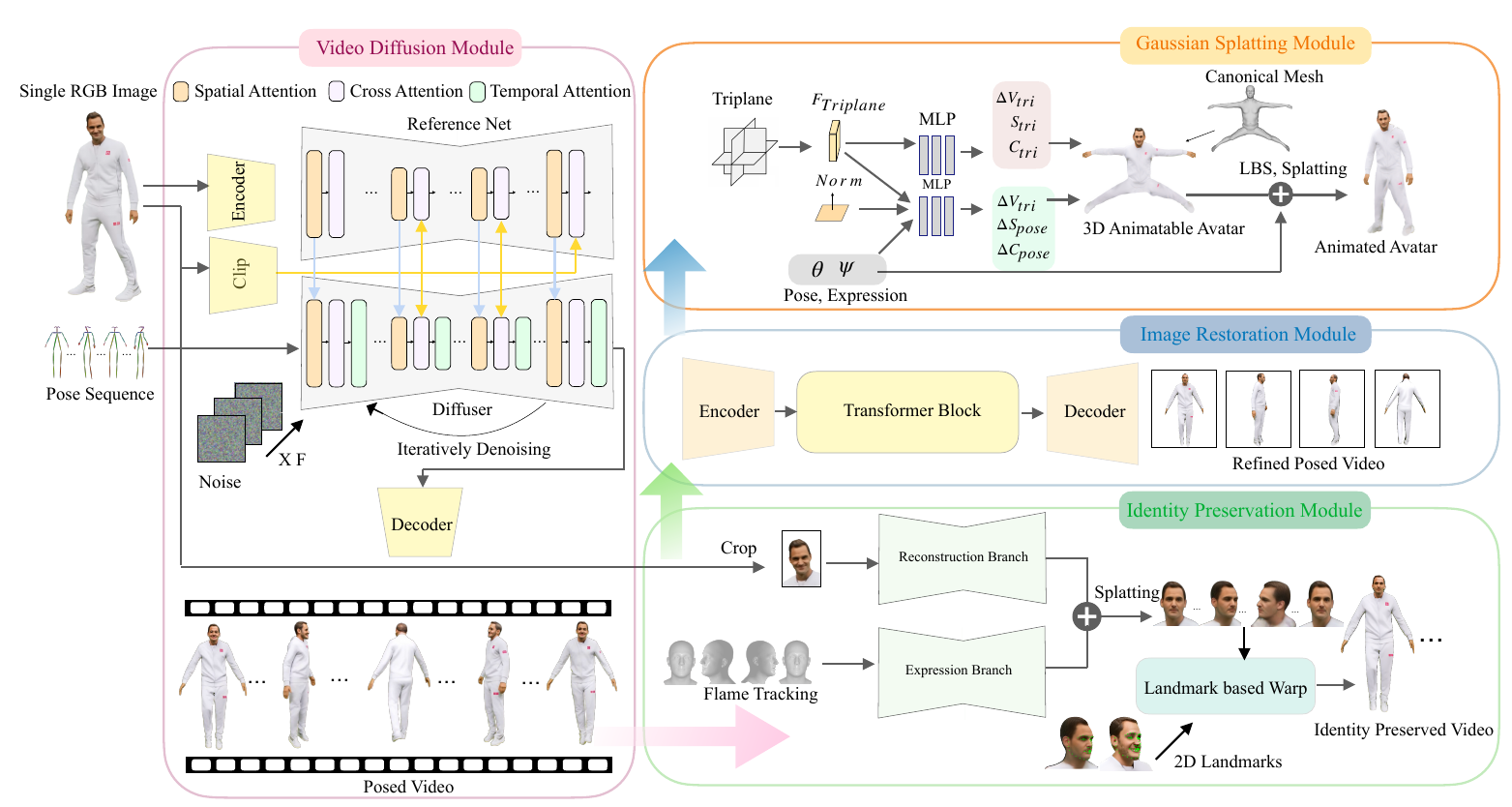}
\caption{\textbf{Overall Pipeline of SVAD.} Starting from a single input image, the diffusion model generates pose-conditioned animations, which are refined using an identity preservation module and an image restoration module. The refined outputs are then used to train the 3DGS avatar, enabling high-fidelity, animatable 3D avatars with consistent details across poses and viewpoints.}
\label{fig:pipeline}
\end{figure*}

\begin{figure*}[ht]
\centering
\includegraphics[width=1.0\linewidth]{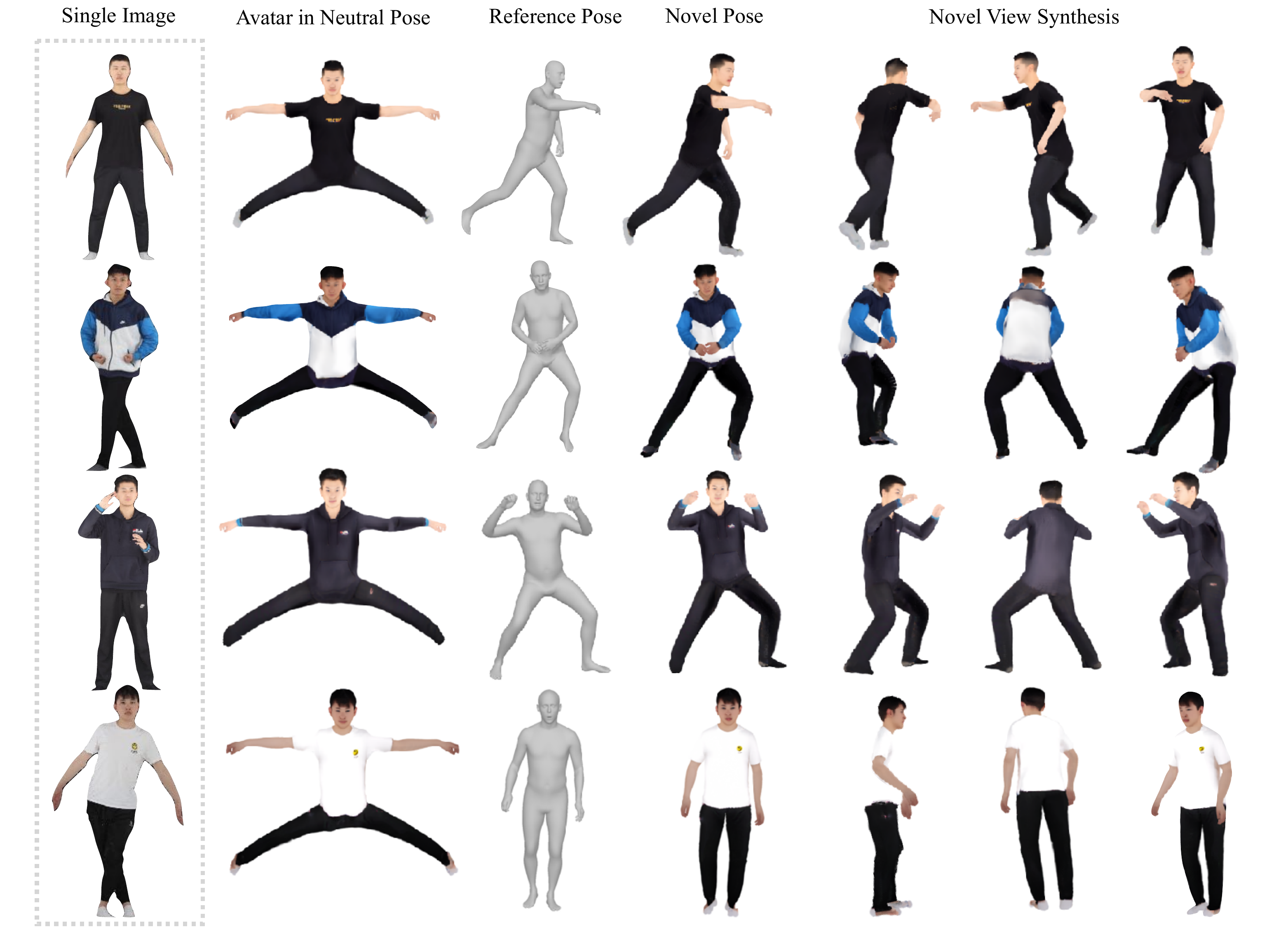}
\caption{\textbf{3D Avatars trained by SVAD.} SVAD generates high quality 3D avatars with just a single-image. The trained avatars can be rendered from any view point, in any pose.}
\label{fig:avatar}
\end{figure*}

\begin{figure*}[ht]
\centering
\includegraphics[width=1.0\linewidth]{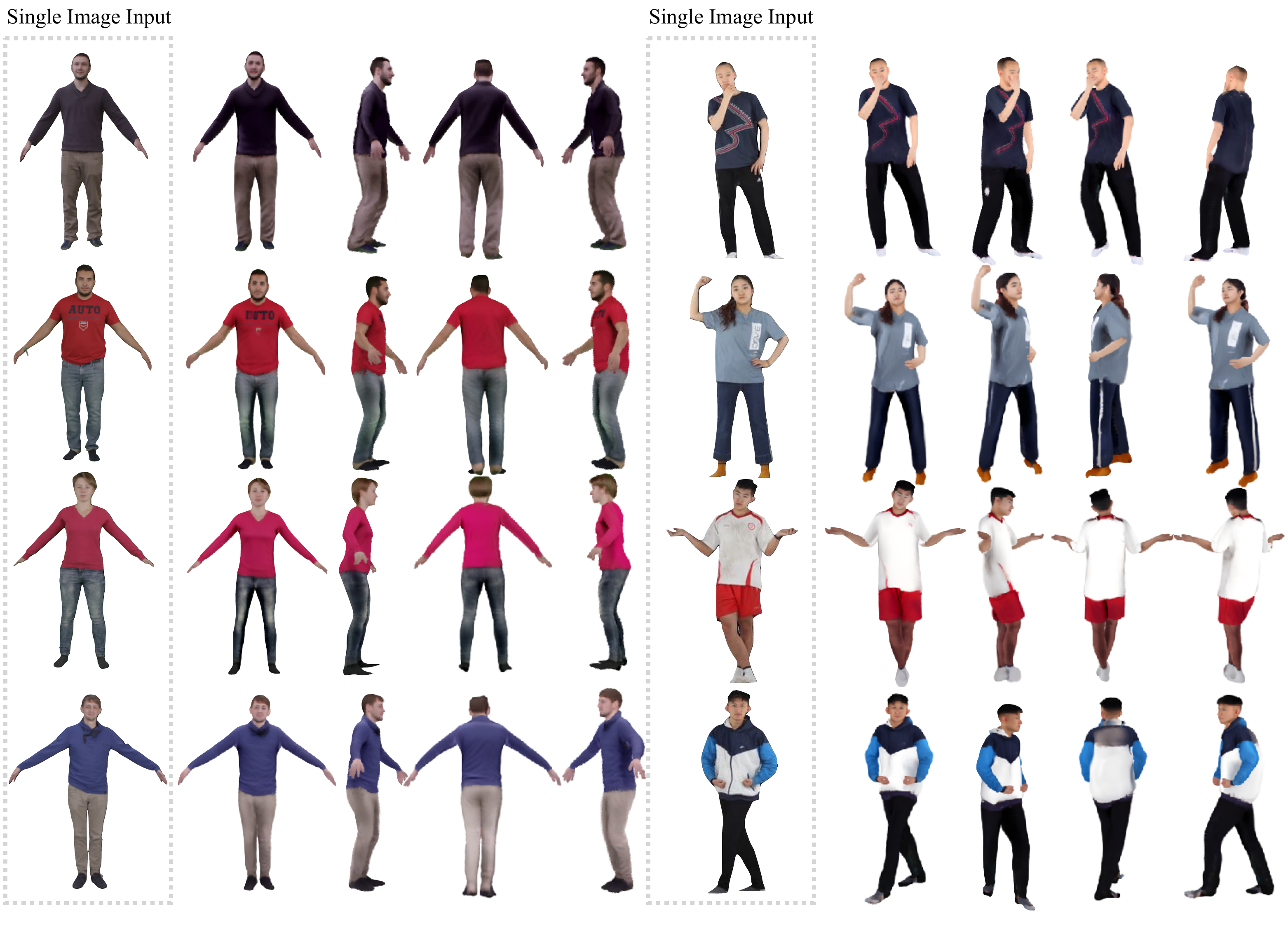}
\caption{\textbf{Qualitative Evaluation} on the People Snapshot dataset and of THuman dataset scan renderings. From a single-image input, SVAD generates high-quality, animatable 3D avatars.}
\label{fig:qual_snapshot}
\end{figure*}

To generate high-quality human avatars from a single-image, facilitating free-viewpoint rendering and realistic animation, we integrate the generative capabilities of video diffusion models with the rendering efficiency of 3D Gaussian-based avatars. We start by leveraging a pretrained video diffusion model~\cite{musepose} for character animation to produce initial synthetic data, as described in Sec.~\ref{subsec:video_diffusion}. Directly using these frames to train a 3DGS avatar model~\cite{moon2024expressive}, however, often yields poor results, with challenges in preserving facial identity, clothing details, and maintaining consistent multi-view coherence across side and back views. To address these issues and enhance avatar quality, we introduce a data augmentation pipeline in Sec.~\ref{subsec:data_augmentation} comprising identity-preservation and image-restoration modules to refine the diffusion outputs. With the augmented synthetic data, we proceed to train a 3DGS avatar model, as outlined in Sec.~\ref{subsec:3dgs_avatar}. The following sections detail the technical methodologies employed in our approach.

\subsection{Video Diffusion Module}
\label{subsec:video_diffusion}
To generate an animated character video \( V \) from a single input image \( I \), we leverage MusePose~\cite{musepose}, a finetuned variant of Animate Anyone~\cite{hu2024animate}, which is a SOTA video diffusion model designed for realistic human animation while maintaining temporal consistency and appearance fidelity. MusePose employs a U-Net~\cite{ronneberger2015u}-based diffusion architecture with integrated pose and temporal controls, allowing for pose-guided animation across frames. For our pipeline, we utilize a pose sequence video from a sequence from the People Snapshot~\cite{alldieck2018video} dataset, which depicts a subject performing a full-body rotation with arms extended horizontally. This sequence results in 189 frames that serve as pose inputs to the MusePose video diffusion model.

The model architecture incorporates several key components for effective character animation. The denoising UNet is implemented as a 3D UNet~\cite{cciccek20163d} with motion modules for temporal coherence. Specifically, we use Vanilla motion modules~\cite{guo2023animatediff,guo2023sparsectrl} with temporal self-attention blocks at resolutions of [1, 2, 4, 8] and in the mid-block. Each transformer~\cite{vaswani2017attention} block contains 8 attention heads, with temporal position encoding enabling positional awareness across a sequence of up to 128 frames. To incorporate pose guidance, a lightweight Pose Guider encodes the motion control signal from the predefined 2D keypoints into a pose-aligned latent representation \( P(p_t) \in \mathbb{R}^{H \times W \times C} \). For a pose feature \( p_t \in \mathbb{R}^{J \times 2} \) at time \( t \), where \( J \) is the number of keypoints, we align the encoding to ensure continuity between frames by adding this encoded pose signal to the noise latent \( z_t \):
\begin{equation}
z_t = z_t + P(p_t)
\end{equation}
For the diffusion process, we adopt a v-prediction~\cite{salimans2022progressive} formulation with zero-SNR sampling~\cite{lin2024common}, using a scaled linear beta schedule with \(\beta_{\text{start}} = 0.00085\) and \(\beta_{\text{end}} = 0.012\). The DDIM~\cite{song2020denoising} sampler is configured for efficient inference with 20 sampling steps and a classifier-free guidance~\cite{ho2022classifier} scale of 3.5.

A critical challenge in character animation is ensuring anatomical consistency between the reference image and the motion poses. Direct application of pose control can result in unnatural animations due to mismatches in body proportions~\cite{chan2019everybody}. Therefore, we employ a comprehensive pose alignment procedure that adapts the source pose to match the reference character's physical characteristics.

Given a reference pose $P_{ref}$ and a source pose $P_{src}$ detected using DWpose~\cite{yang2023effective}, we compute scale parameters $\mathbf{S} = \{s_1, s_2, \ldots, s_{10}\}$ for ten distinct body regions: neck, face, shoulders, upper arms, lower arms, hands, torso, upper legs, and lower legs. For each body part $i$, we compute its scale factor $s_i$ as the ratio between the corresponding keypoint distances. For body parts with bilateral symmetry (e.g., arms), we average the scales from both sides:
\begin{equation}
s_{arm\_upper} = \frac{1}{2}\left(\frac{\|p_{ref}^2 - p_{ref}^3\|}{\|p_{src}^2 - p_{src}^3\|} + \frac{\|p_{ref}^5 - p_{ref}^6\|}{\|p_{src}^5 - p_{src}^6\|}\right)
\end{equation}
To apply these scales to the source pose, we use a rotation matrix transformation centered at anchor points specific to each body part:
\begin{equation}
p' = c_i + s_i \cdot (p - c_i)
\end{equation}
where $c_i$ is the anchor center for part $i$. This hierarchical approach ensures body proportions match the reference while maintaining the overall pose structure.

\subsection{Data Augmentation Module}
\label{subsec:data_augmentation}
Training the 3DGS model using only outputs from the video diffusion model often results in low-fidelity avatars, particularly in terms of facial details and high-frequency features like hands and clothing. To address these challenges, we introduce a data augmentation module that enhances the quality of the training data. This module includes an identity preservation sub-module ensuring coherence in facial details across frames and a image restoration submodule which refines texture quality and high-frequency details, resulting in more realistic textures. This comprehensive data augmentation significantly improves the synthetic training data, enabling the 3DGS avatar model integrated in the future to generate more realistic and detailed 3D avatars.

\noindent\textbf{Identity preservation sub-module.}
\noindent To ensure consistent and realistic facial details across frames, we implement an identity preservation module that combines 3D head reconstruction and facial fusion techniques. From a single input image, we first create a 3D Gaussian-based head avatar using a method inspired by Chu \etal~\cite{chu2024generalizable}, which employs a novel \textit{dual-lifting} approach that predicts both forward and backward lifting distances.

Given an input image \( I_s \), global and local features \( F_{\text{local}} \) are extracted using a frozen DINOv2~\cite{oquab2023dinov2} backbone. These features are used to predict forward and backward lifting distances, positioning 3D Gaussians \( G_{\text{pos}} \) as follows:
\begin{equation}
G_{\text{pos}} = [\mathbf{p}_s + E_{\text{Conv0}}(F_{\text{local}}) \cdot \mathbf{n}_s, \mathbf{p}_s - E_{\text{Conv1}}(F_{\text{local}}) \cdot \mathbf{n}_s],
\end{equation}
\noindent where \( \mathbf{p}_s \) is the initial point plane, \( \mathbf{n}_s \) is the normal vector, and \( E_{\text{Conv}} \) are convolutional layers predicting offsets. To capture expression variations, we bind 3DMM~\cite{li2017learning} features: 
\begin{equation}
G_{\text{expr}} = \text{MLP}(F_{\text{3DMM}} + F_{\text{global}}).
\end{equation}

To animate this 3D head avatar, we separately track FLAME~\cite{li2017learning} parameters \(\Theta = \{\beta, \psi, \theta, \phi\}\) from our predefined pose sequence video (the same sequence used in the video diffusion module), where \(\beta \in \mathbb{R}^{300}\) represents shape parameters, \(\psi \in \mathbb{R}^{100}\) expression parameters, \(\theta \in \mathbb{R}^{6}\) global pose parameters, and \(\phi \in \mathbb{R}^{6}\) eye pose parameters. These tracked parameters serve as animation controls for the reconstructed 3D head. Using these tracked FLAME parameters, we render the 3D head avatar to generate a sequence of head images that match our predefined pose sequence. These renderings provide high-quality, identity-consistent facial details across different viewpoints. Since the quality of the renderings deteriorates for back-of-head views, we selectively apply the face fusion process only to frames where the head is front-facing (front and side views).

For the face fusion process, we detect facial landmarks~\cite{dlib} on both the diffusion-generated frame \( I_{\text{orig}} \) and the rendered head image \( I_{\text{head}} \), compute an affine transformation for alignment, and use Poisson image editing~\cite{perez2023poisson} for seamless blending:
\begin{equation}
\min_{I} \int_{\Omega} \left\| \nabla I - \nabla I_{\text{warp}} \right\|^2 \, \mathrm{d}x \, \mathrm{d}y, \quad \text{subject to } I|_{\partial \Omega} = I_{\text{orig}}|_{\partial \Omega},
\end{equation}
where \( \Omega \) is defined by the facial mask. This ensures temporally consistent facial details while preserving the original identity throughout the animation sequence.

\noindent\textbf{Image restoration sub-module.}
\noindent Finally, to preserve quality of fine detailed regions, we employ an image restoration module based on the work of Chen \etal~\cite{chen2024towards}, specifically their diffusion-based image restoration method BFRffusion. This approach leverages the generative prior encapsulated in the pretrained Stable Diffusion~\cite{rombach2022high} model to enhance image details through a comprehensive architecture that effectively extracts features from low-quality images and restores realistic facial details.

For our implementation, we set the super-resolution scale factor to $s=1.5$, which our empirical analysis showed provides an optimal balance between detail enhancement and artifact suppression. We observed that scale factors $s < 1.5$ produce insufficient detail recovery, while factors $s > 2.0$ introduce perceptual artifacts (particularly in specular regions such as eyes) and significantly increase computational demands during avatar training. The diffusion process uses 50 DDIM sampling steps with:
\begin{equation}
z_{t-1} = \sqrt{\alpha_{t-1}}\left(\frac{z_t - \sqrt{1-\alpha_t}\epsilon_\theta(z_t)}{\sqrt{\alpha_t}}\right) + \sqrt{1-\alpha_{t-1}}\epsilon_\theta(z_t)
\end{equation}
where $\alpha_t = \prod_{i=1}^t (1-\beta_i)$ and $\epsilon_\theta$ is the denoising network.
We utilize a classifier-free guidance scale of $w=3.5$, with the guidance equation:
\begin{equation}
\hat{\epsilon}_\theta(z_t) = (1 + w)\epsilon_\theta(z_t) - w\epsilon_\theta(z_t, \emptyset)
\end{equation}
where $\epsilon_\theta(z_t, \emptyset)$ represents the unconditional prediction. This achieves an optimal balance between restoration quality and processing speed. For face regions, the method employs a face restoration helper with facial landmark detection to specifically enhance facial details, ensuring identity consistency across generated frames. Restored faces are blended with Poisson image editing.

This image restoration submodule significantly improves the fidelity and realism of our synthetic training data by restoring fine facial details, enhancing texture quality in clothing and accessories, and improving overall image coherence. The refined data enables the 3DGS avatar to learn more accurate representations with consistent high-frequency details that persist across poses and viewpoints.

\subsection{3D Human Gaussian Splatting Module}
\label{subsec:3dgs_avatar}
We apply the architecture of a 3DGS based avatar method introduced by Moon \etal \cite{moon2024expressive}, which integrates the SMPL-X~\cite{pavlakos2019expressive} model with a 3D Gaussian-based representation to produce animatable human avatars. Each 3D Gaussian acts as a vertex connected by a pre-defined mesh topology following SMPL-X. This hybrid representation combines the expressive surface modeling of SMPL-X with the flexibility of a volumetric approach, allowing for smooth interpolation across the body surface essential for realistic animations.

Each Gaussian point is associated with positional data \( \mathbf{V} \in \mathbb{R}^{N \times 3} \), RGB color values \( \mathbf{C} \in \mathbb{R}^{N \times 3} \), and a scale parameter \( \mathbf{S} \in \mathbb{R}^{N} \), where \( N \) is the number of Gaussians. The Gaussian splatting rendering equation is:
\begin{equation}
I = f(V, \exp(S), C, K, E),
\end{equation}
where \( V \) represents positions, \( S \) denotes scale, \( C \) colors, and \( K \) and \( E \) camera parameters. 

Pose-dependent deformations are applied through an MLP network, predicting offsets for each Gaussian based on SMPL-X pose parameters:
\begin{equation}
\mathbf{V}_{\text{pose}} = \mathbf{V} + \Delta \mathbf{V}_{\text{pose}} + \Delta \mathbf{V}_{\text{expr}}.
\end{equation}
To maintain spatial coherence, a Laplacian regularizer~\cite{nealen2006laplacian,sorkine2004laplacian} minimizes the difference between the Laplacian of the canonical mesh and the deformed Gaussian points:
\begin{equation}
L_{\text{Lap}} = \left\| \Delta \mathbf{V}_{\text{canonical}} - \Delta \mathbf{V}_{\text{deformed}} \right\|^2.
\end{equation}
This approach combined with our augmented synthetic data achieves highly realistic, animatable avatars capable of real-time rendering with smooth deformations across facial expressions, body movements, and hand gestures.

\input{tables/quant_snapshot}

%% file: tables/quant_snapshot.tex
\begin{table*}[t]
\centering
\setlength{\tabcolsep}{4pt}
\resizebox{\textwidth}{!}{
\begin{tabular}{lcccccccccccc}
\toprule
\multirow{2}{*}{Method} & \multicolumn{3}{c}{Female-4-casual} & \multicolumn{3}{c}{Male-3-casual} & \multicolumn{3}{c}{Female-3-casual} & \multicolumn{3}{c}{Male-4-casual} \\
\cmidrule(lr){2-4} \cmidrule(lr){5-7} \cmidrule(lr){8-10} \cmidrule(lr){11-13}
 & PSNR↑ & SSIM↑ & LPIPS↓ & PSNR↑ & SSIM↑ & LPIPS↓ & PSNR↑ & SSIM↑ & LPIPS↓ & PSNR↑ & SSIM↑ & LPIPS↓ \\
\midrule
HumanNeRF~\cite{weng2022humannerf} & 27.07 & 0.9615 & 0.0151 & 26.90 & 0.9605 & 0.0181 & 24.46 & 0.9516 & 0.0269 & 25.50 & 0.9397 & 0.0357 \\
GaussianAvatar~\cite{hu2024gaussianavatar} & 30.84 & 0.9771 & \cellcolor{strongblue}0.0140 & \cellcolor{strongblue}30.98 & \cellcolor{strongblue}0.9790 & \cellcolor{strongblue}0.0145 & 29.55 & \cellcolor{strongblue}0.9762 & \cellcolor{strongblue}0.0220 & 28.78 & \cellcolor{strongblue}0.9755 & \cellcolor{strongblue}0.0230 \\
ExAvatar~\cite{moon2024expressive} & \cellcolor{strongblue}30.98 & \cellcolor{strongblue}0.9789 & 0.0333 & 29.75 & 0.9628 & 0.0402 & \cellcolor{strongblue}29.74 & 0.9678 & 0.0458 & \cellcolor{strongblue}28.89 & 0.9666 & 0.0500 \\
ExAvatar~\cite{moon2024expressive} (Single Image)& \cellcolor{lightred}20.42 & \cellcolor{lightred}0.9427 & \cellcolor{lightred}0.0656 & \cellcolor{strongred}23.24 & \cellcolor{lightred}0.9448 & \cellcolor{lightred}0.0562 & \cellcolor{lightred}20.12 & \cellcolor{lightred}0.9492 & \cellcolor{lightred}0.0543 & \cellcolor{strongred}23.74 & \cellcolor{lightred}0.9497 & \cellcolor{lightred}0.0610 \\
Ours (Single Image) & \cellcolor{strongred}21.51 & \cellcolor{strongred}0.9442 & \cellcolor{strongred}0.0528 & \cellcolor{lightred}22.54 & \cellcolor{strongred}0.9467 & \cellcolor{strongred}0.0484 & \cellcolor{strongred}21.96 & \cellcolor{strongred}0.9609 & \cellcolor{strongred}0.0541 & \cellcolor{lightred}23.71 & \cellcolor{strongred}0.9570 & \cellcolor{strongred}0.0592 \\
\bottomrule
\end{tabular}
}
\caption{\textbf{Quantitative Evaluation} on the People Snapshot~\cite{alldieck2018video} Dataset. Our approach demonstrates superior performance on \textit{single-image} input, outperforming the baseline on most of the metrics. The top two results for \textit{single-image} input are highlighted in \colorbox{strongred}{first} and \colorbox{lightred}{second}, with the overall best result highlighted in \colorbox{strongblue}{first}. Note that methods that use monocular input utilize approximately 200 input frames.}
\label{tab:quant_snapshot}
\end{table*}

%% file: sec/4_experiments.tex
\section{Experiments}
\label{sec:experiments}
\subsection{Datasets and Metrics}
\label{subsec:datasets_metrics}

\noindent\textbf{People-Snapshot Dataset}~\cite{alldieck2018video} We conduct our avatar evaluation on the People-Snapshot dataset, which features video recordings of subjects performing 360-degree rotations. Following both Anim-NeRF~\cite{weng2022humannerf} and InstantAvatar~\cite{jiang2023instantavatar}, we address a known limitation in this dataset: the provided pose parameters often exhibit misalignment with the actual image content. Anim-NeRF addressed this by optimizing pose parameters for both training and test sequences. To ensure fair comparison with existing methods, we adopt these same optimized pose parameters and keep them frozen throughout our training process for fair comparison.

\noindent\textbf{THuman Dataset}~\cite{tao2021function4d} For evaluating single-image 3D human reconstruction, we employ the THuman dataset, adhering to the methodology established in Ultraman~\cite{chen2024ultraman}. Our procedure involves randomly selecting 100 scans and generating renderings from four viewpoints (front, left, right, back). We then measure the similarity between our reconstructed outputs and the ground-truth scan renderings from these identical perspectives, facilitating objective comparison with other SOTA methods.

\noindent\textbf{Evaluation Metrics} Our evaluation framework uses four metrics to quantify reconstruction quality: PSNR~\cite{gonzalez2008digital}, SSIM~\cite{wang2004image}, LPIPS~\cite{zhang2018unreasonable}, and CLIP Similarity~\cite{radford2021learning} (referred to as CLIP in our tables). This combination provides comprehensive assessment across different dimensions: PSNR for pixel accuracy, SSIM for structural coherence, LPIPS for perceptual alignment with human vision, and CLIP for semantic consistency at the feature level. The use of these metrics enables thorough evaluation of both fine-grained detail, and overall perceptual quality.

\subsection{Quantitative Evaluation}

We quantitatively evaluate the quality of single-image 3D avatars generated by our method against SOTA 3D avatar generation methods~\cite{weng2022humannerf,hu2024gaussianavatar,moon2024expressive}. While current 3D avatar models generally require a monocular video as input, we assess our model’s performance using a single-image as input on ExAvatar~\cite{moon2024expressive}. Additionally, we report results using the original full training set of approximately 200 input frames for monocular input based avatar models for reference. As shown in Table~\ref{tab:quant_snapshot}, our model achieves highest scores on most of the metrics among single-image input methods.
We further compare our approach with single-view 3D human reconstruction methods~\cite{saito2019pifu,chen2024ultraman,zhang2024sifu,ho2024sith,huang2024tech}, many of which employ the SMPL model, allowing for animatability through mesh fitting and reposing techniques, such as those in Editable Humans~\cite{ho2023learning}. We randomly sample 100 scans from the THuman dataset and report results. We repose our trained avatar using ground-truth SMPL-X parameters and compare with the ground-truth scan renderings from the same views. As presented in Table~\ref{tab:quant_thuman_mesh}, our method surpasses all baselines, demonstrating superior quality in 3D human reconstruction tasks.

\subsection{Qualitative Evaluation}

\begin{figure}[t]
\includegraphics[width=1.0\columnwidth, trim={0.6cm 0cm 0.2cm 0cm}]{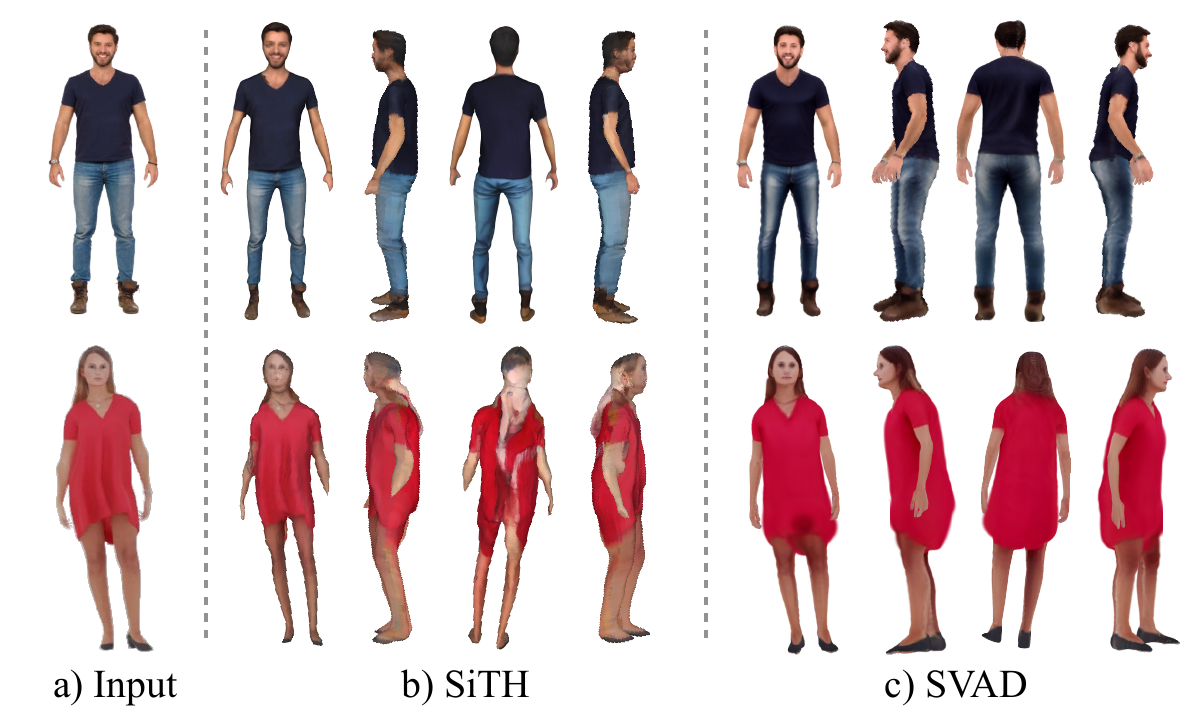}
\caption{\textbf{Qualitative Evaluation} against SiTH~\cite{ho2024sith}. Our approach better reconstructs complex contours and subtle features, resulting in a more lifelike and coherent side-view appearance.}
\label{fig:qual_sith}
\end{figure}

\begin{figure}[t]
\includegraphics[width=1.0\columnwidth, trim={0.6cm 0cm 0.2cm 0cm}]{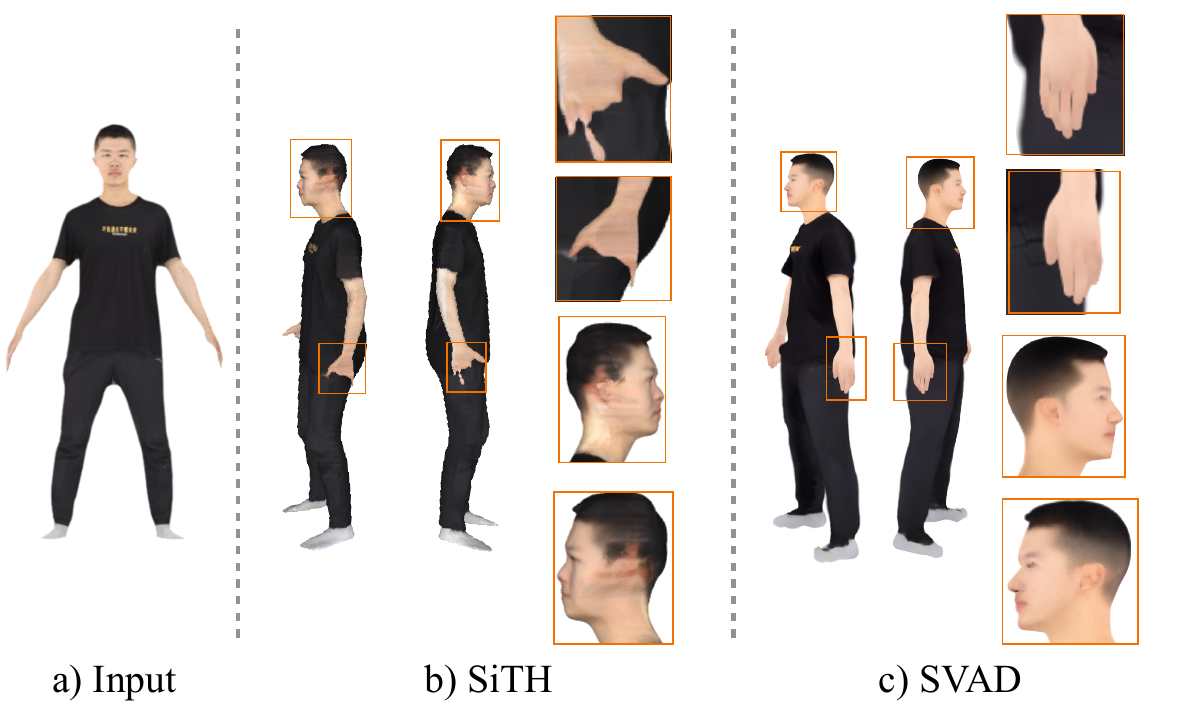}
\caption{\textbf{Qualitative Evaluation} against SiTH~\cite{ho2024sith}. Our method reconstructs fine detail (hands), while preserving original identity in facial regions.}
\label{fig:qual_sith_thuman}
\vspace{-10pt}
\end{figure}

\begin{figure}[t]
\includegraphics[width=1.0\columnwidth, trim={0.6cm 0cm 0.2cm 0cm}]{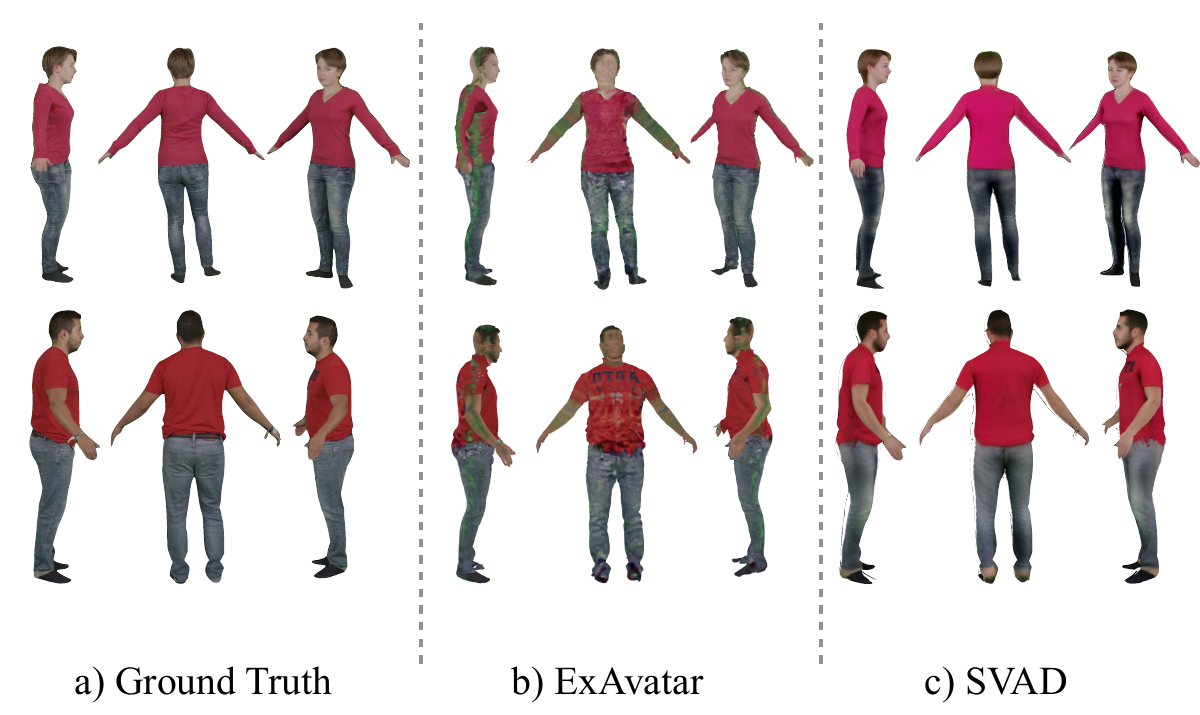}
\caption{\textbf{Qualitative Evaluation} against ExAvatar~\cite{moon2024expressive} in single-image to 3D avatar task. Our method generates more plausible back and side views with the generated synthetic dataset.}
\label{fig:qual_exavatar}
\vspace{-10pt}
\end{figure}

Figure~\ref{fig:qual_snapshot} shows the overall quality of our generated 3D avatars from single-images in the People Snapshot and the THuman dataset. Figure~\ref{fig:qual_sith}, Figure~\ref{fig:qual_sith_thuman} shows that our method performs superior compared to current SiTH~\cite{ho2024sith}. For single-image avatar generation, we evaluate on the People Snapshot dataset and compare against ExAvatar~\cite{moon2024expressive}. For fairness, we train ExAvatar for the same number (12,000) of iterations. Figure ~\ref{fig:qual_exavatar} shows that for single-image avatar generation, our method performs superior especially for the back and side views. 
 
\input{tables/quant_thuman_mesh}
\input{tables/ablation_snapshot}
\input{tables/ablation_thuman}

\begin{figure}[t]
\includegraphics[width=1.0\columnwidth, trim={0.6cm 0cm 0.2cm 0cm}]{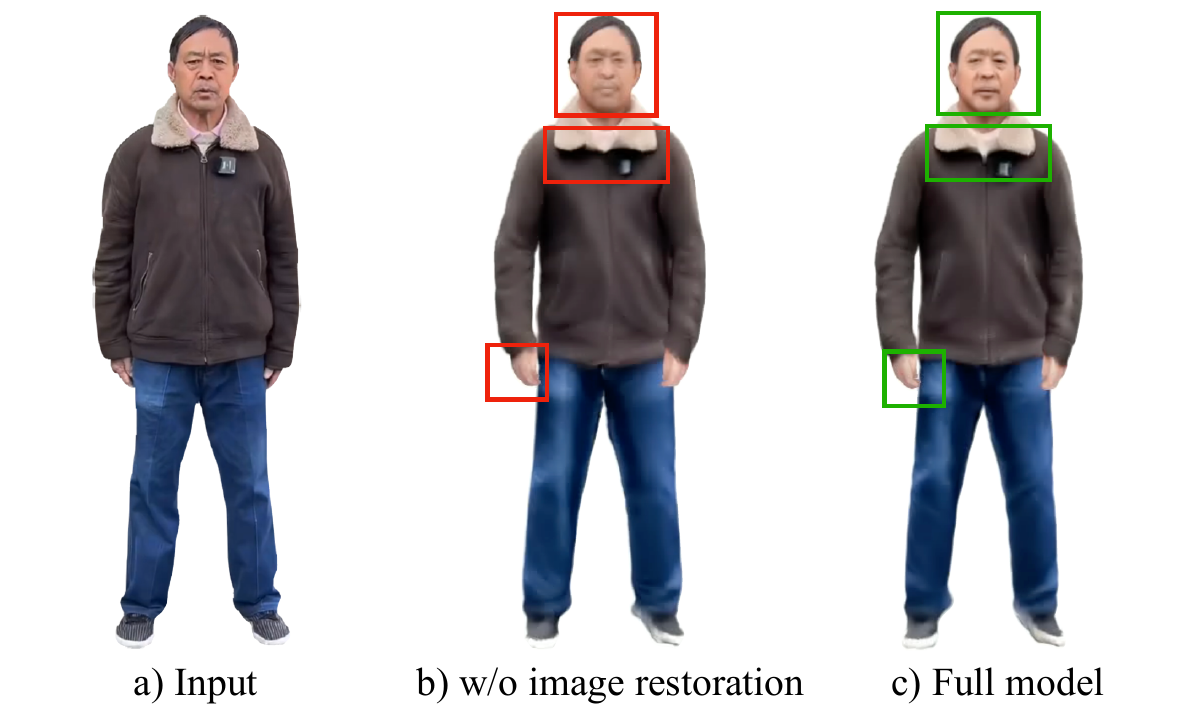}
\caption{\textbf{Ablation study} on the image restoration module. We show that applying the module into our pipeline recover fine details on the final avatar output.}
\label{fig:ab_restor}
\vspace{-10pt}
\end{figure}

\begin{figure}[t]
\includegraphics[width=1.0\columnwidth, trim={0.6cm 0cm 0.2cm 0cm}]{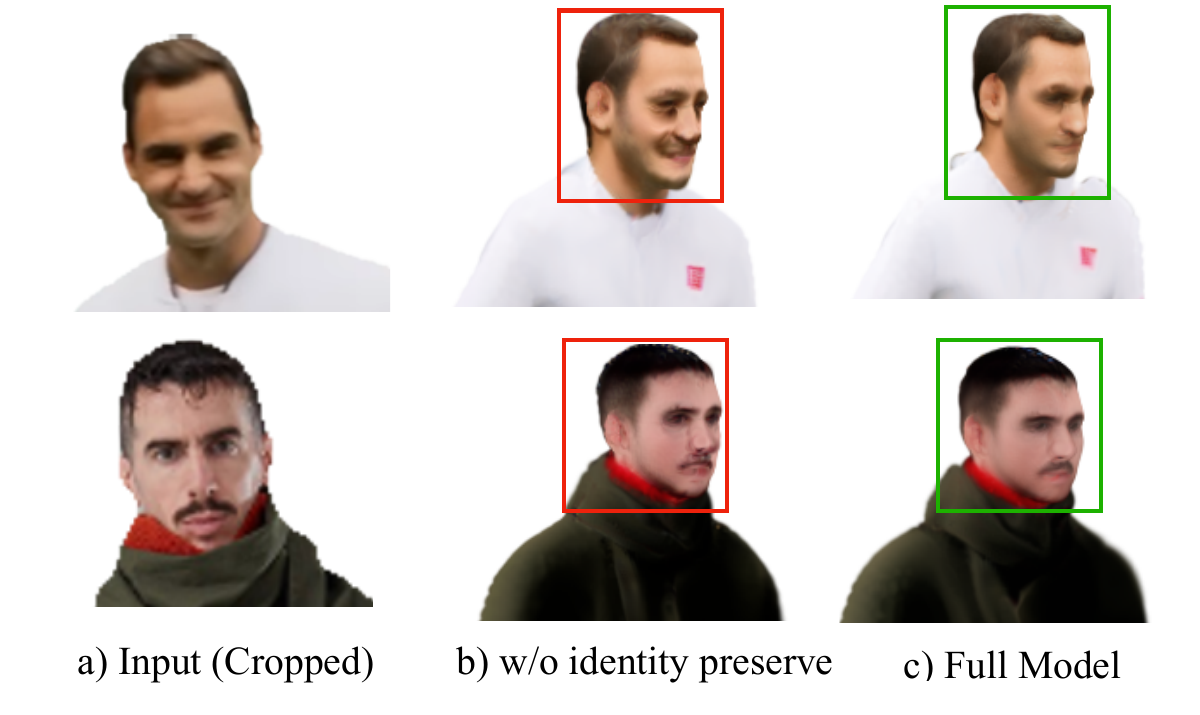}
\caption{\textbf{Ablation study} on the identity preservation module. We show that with the module, the final avatar maintains facial details on the original input image.}
\label{fig:ab_face}
\vspace{-10pt}
\end{figure}

\subsection{Ablation Study}
In this section, we conduct ablation studies to validate each component of our methods. The average metrics over 4 sequences in the People Snapshot dataset are reported in Table~\ref{tab:ablation_people_snapshot}. It shows that our methods modules are required to reach the optimal performance reflected by all the metrics. Using the THuman dataset, we apply the same evaluation technique as in our quantitative evaluation. Results show that our method performs the best in PSNR, SSIM and CLIP similarity and performs second best in LPIPS. Figure~\ref{fig:ab_restor} shows visual results of the effect of the image restoration module. High-detailed regions such as clothing texture, fingers, and facial details are better preserved when applying our module. Figure~\ref{fig:ab_face}, shows the visual effect of the identity preservation module. We clearly show that original input's facial details are more preserved our module.  

%% file: tables/quant_thuman_mesh.tex
\begin{table}[t]
\setlength{\tabcolsep}{6pt}  
\renewcommand{\arraystretch}{1.1}  
\small  
\centering  
\begin{tabular}{l@{\hskip 6pt}cccc}  
\toprule
\textbf{Method} & \textbf{PSNR$\uparrow$} & \textbf{SSIM$\uparrow$} & \textbf{LPIPS$\downarrow$} & \textbf{CLIP$\uparrow$} \\
\midrule
PIFu~\cite{saito2019pifu}  & 15.62 & 0.8921 & 0.1903 & 0.8612 \\
TeCH~\cite{huang2024tech} & 15.85 & 0.8892 & 0.1667 & 0.8890 \\
Ultraman~\cite{chen2024ultraman}  & 18.13 & \cellcolor{lightred}0.9019 & 0.1334 & \cellcolor{lightred}0.9089 \\
SIFU~\cite{zhang2024sifu} & 18.59 & 0.8591 & 0.1402 & 0.8873 \\
SiTH~\cite{ho2024sith}  & \cellcolor{lightred}19.98 & 0.9018 & \cellcolor{lightred}0.1294 & 0.9084 \\
\textbf{Ours} & \cellcolor{strongred}\textbf{20.92} & \cellcolor{strongred}\textbf{0.9291} & \cellcolor{strongred}\textbf{0.1124} & \cellcolor{strongred}\textbf{0.9321} \\
\bottomrule
\end{tabular}
\caption{\textbf{Quantitative Evaluation} on single-image to 3D human reconstruction tasks on 100 scan renderings of the THuman~\cite{tao2021function4d} Dataset. Top two results are colored as \colorbox{strongred}{first} \colorbox{lightred}{second}.}
\label{tab:quant_thuman_mesh}
\end{table}

%% file: tables/ablation_snapshot.tex
\begin{table}[t]
\centering
\setlength{\tabcolsep}{6pt}
\resizebox{\columnwidth}{!}{ 
\begin{tabular}{lcccc}
\toprule
\textbf{Method} & \textbf{PSNR}$\uparrow$ & \textbf{SSIM}$\uparrow$ & \textbf{LPIPS}$\downarrow$ & \textbf{CLIP}$\uparrow$ \\
\midrule
w/o Identity Preserve & 22.19 & \cellcolor{lightred}0.9419 & \cellcolor{lightred}0.0623 & \cellcolor{lightred}0.9231 \\
w/o Image Restoration & \cellcolor{lightred}22.61 & 0.9298 & 0.0645 & 0.9239 \\
\textbf{Ours (Full)} & \cellcolor{strongred}22.79 & \cellcolor{strongred}0.9502 & \cellcolor{strongred}0.0594 & \cellcolor{strongred}0.9241 \\
\bottomrule
\end{tabular}
}
\caption{\textbf{Ablation study} on the People Snapshot dataset. Our full model consistently outperforms variants with individual components removed across all metrics.}
\label{tab:ablation_people_snapshot}
\end{table}

%% file: tables/ablation_thuman.tex
\begin{table}[t]
\centering
\setlength{\tabcolsep}{6pt}
\resizebox{\columnwidth}{!}{ 
\begin{tabular}{lcccc}
\toprule
\textbf{Method} & \textbf{PSNR}$\uparrow$ & \textbf{SSIM}$\uparrow$ & \textbf{LPIPS}$\downarrow$ & \textbf{CLIP}$\uparrow$ \\
\midrule
w/o Identity Preserve & 20.12 & \cellcolor{lightred}0.9256 & 0.1294 & \cellcolor{lightred}0.9284 \\
w/o Image Restoration & \cellcolor{lightred}20.16 & 0.9212 & \cellcolor{strongred}0.0799 & 0.9201 \\
\textbf{Ours (Full)} & \cellcolor{strongred}20.92 & \cellcolor{strongred}0.9291 & \cellcolor{lightred}0.1124 & \cellcolor{strongred}0.9321 \\
\bottomrule
\end{tabular}
}
\caption{\textbf{Ablation study} on the THuman dataset. The full model achieves superior performance in most metrics, demonstrating the importance of each component in our pipeline.}
\label{tab:ablation_thuman}
\vspace{-10pt}
\end{table}

%% file: sec/5_conclusion.tex
\section{Conclusion and Discussion}
\label{sec}
In this work, we introduced SVAD, a novel synthetic data generation approach for creating high-fidelity, animatable 3D human avatars from a single image. By combining the generative power of diffusion models with the rendering efficiency of 3D Gaussian Splatting, SVAD produces avatars that maintain consistent identity across varied poses and viewpoints. Through comprehensive experiments, we demonstrate that our method achieves SOTA performance.

\noindent\textbf{Limitations and Future Work.}
Our method faces several limitations. First, inaccurate background segmentation of training frames produces floating artifacts. Second, our approach struggles with complex clothing textures and loose outfits due to limitations of the video diffusion model in generating detailed synthetic data. Finally, the computational requirements present practical challenges—the video diffusion step demands substantial resources, and the complete pipeline requires 5-6 hours per avatar generation. Future work will focus on improving handling of diverse clothing types and optimizing computational performance.

%% file: sec/X_suppl.tex
\clearpage
\section{Implementation Details}
\label{supple-sec1}
In this section, we provide comprehensive technical details of SVAD. We first describe the predefined pose sequences that serve as conditioning inputs for our video diffusion model. Next, we elaborate on the video diffusion module, the identity preservation module and image restoration module for enhancing facial fidelity and overall texture quality. Finally, we elaborate on the training process for our 3DGS avatar, including the SMPL-X~\cite{pavlakos2019expressive} parameter fitting procedure and the optimization strategy for the 3D Gaussian representation.

\subsection{Predefined Pose Sequences} 
To initialize frame generation for our pipeline, we rely on a predefined set of poses extracted from the People Snapshot~\cite{alldieck2018video} dataset. Specifically, we utilize the \textit{male-4-casual} sequence, which depicts a subject performing a full-body rotation with arms extended horizontally. Using DWPose~\cite{yang2023effective}, we extract 2D keypoints \( K \in \mathbb{R}^{J \times 2} \), where \( J = 17 \) is the number of keypoints, from this sequence to create a standardized pose template. This sequence serves as the conditioning input for the video diffusion model, resulting in $189$ frames of pose-guided human animation, with a resolution of \( 1024 \times 1024 \).

Our experiments revealed that inference with lower resolutions such as \( 512 \times 512 \) produced animations with significantly degraded facial details, which adversely affected subsequent processing steps. Particularly, the landmark-based face fusion technique requires accurate facial landmark detection, which proved unreliable on low-resolution outputs. The absence of distinct facial features in \( 512 \times 512 \) outputs led to inconsistent landmark detection, compromising the accuracy of 3D head rendering and warping operations. The higher \( 1024 \times 1024 \) resolution preserves critical facial details, enabling robust landmark detection and consistent face fusion results across the generated sequence.

\subsection{Video Diffusion Module}
For our video diffusion module, we leverage MusePose~\cite{musepose}, a modified variant of Animate Anyone~\cite{hu2024animate}, specifically designed for pose-guided video generation from a single image. The architecture follows a UNet-based~\cite{ronneberger2015u} denoising diffusion model with temporal modeling capabilities, enabling coherent video generation while maintaining consistency with the reference image.

During inference, the video diffusion pipeline performs iterative denoising of random noise guided by the reference image and pose sequence. We configure the DDIM sampler~\cite{song2020denoising} with 20 sampling steps and a classifier-free guidance~\cite{ho2022classifier} scale of 3.5 which keeps balance between generation quality and inference speed. The network architecture employs a 3D variant of the standard UNet architecture, where temporal layers enable information exchange across video frames. The reference image features are extracted using a CLIP vision encoder~\cite{radford2021learning} and processed through a reference UNet. These features are transferred to the denoising UNet via a custom attention mechanism:
\begin{equation}
\text{Attn}(Q, K, V) = \text{softmax}\left(\frac{QK^T}{\sqrt{d}}\right)V
\end{equation}
where $Q$ represents queries from the denoising UNet features, while $K$ and $V$ are derived from the reference image features. This mechanism ensures that generated frames maintain the appearance details of the reference image.

The pose conditioning is handled by the PoseGuider module, which processes pose skeleton images through a series of convolutional layers to create pose feature embeddings. These embeddings are added to the latent noise to spatially align the generation with target poses:
\begin{equation}
z_t = z_t + P(p_t)
\end{equation}
where $z_t$ is the noise latent at timestep $t$, $p_t \in \mathbb{R}^{J \times 2}$ is the pose feature at time $t$, and $P(\cdot)$ represents the pose guider. The PoseGuider has an input convolutional layer, followed by blocks with increasing channel dimensions $(16, 32, 64, 128)$, and a zero-initialized output projection to the conditioning embedding channels.

For handling longer video sequences beyond the model's context window, we employ a sliding window~\cite{ho2022video} approach. The model processes frames in overlapping chunks of length $S=48$ with an overlap of $O=4$ frames. This enables the generation of arbitrarily long sequences while maintaining temporal consistency. The generative process for each video segment can be expressed as:
\begin{equation}
V_{i:i+S} = \mathcal{G}(I_\text{ref}, P_{i:i+S}, z)
\end{equation}
where $V_{i:i+S}$ represents the generated video segment from frame $i$ to $i+S$, $\mathcal{G}$ is our diffusion model, $I_\text{ref}$ is the reference image, $P_{i:i+S}$ are the corresponding pose skeletons, and $z$ is the random noise. By processing these overlapping segments and blending them at the boundaries, the final full-length human-animated video has smooth transitions.

\subsection{Identity Preservation Module} 
Following the initial frame generation by the video diffusion model, we refine the facial regions to enhance identity consistency and detail preservation. Our identity preservation pipeline consists of three main components: FLAME~\cite{li2017learning} parameter tracking~\ref{flame_tracking}, 3D head rendering~\ref{head_rendering}, and face fusion~\ref{face_fusion}. Each component plays a crucial role in generating high-quality, identity-consistent facial regions in our data augmentation pipeline.

\subsubsection{FLAME Parameter Tracking}
\label{flame_tracking}
We begin by tracking FLAME parameters from our predefined pose sequence video to guide the animation of our 3D head avatar. Using a tracking engine with focal length set to $12.0$, we extract parameters $\Theta = \{\beta, \psi, \theta, \phi\}$, where $\beta \in \mathbb{R}^{300}$ represents shape parameters, $\psi \in \mathbb{R}^{100}$ expression parameters, $\theta \in \mathbb{R}^{6}$ global pose parameters, and $\phi \in \mathbb{R}^{6}$ eye pose parameters.

To ensure smooth parameter transitions across frames, we apply Savitzky-Golay~\cite{john2021adaptive} filtering with a window length of $9$ frames and polynomial order of $2$. For rotation parameters, we employ quaternion-based smoothing~\cite{zhao2017path} with a continuity enforcement algorithm to handle sign flips:

\begin{equation}
\text{q}'_{t+1} = 
\begin{cases}
-\text{q}_{t+1}, & \text{if } \text{q}_t \cdot \text{q}_{t+1} < 0 \\
\text{q}_{t+1}, & \text{otherwise}
\end{cases}
\end{equation}
Different parameter types are smoothed with specific momentum coefficients: rotation matrices $\alpha=0.6$, translation vectors $\alpha=0.6$, and eye pose parameters $\alpha=0.7$. This comprehensive smoothing strategy eliminates jitter and ensures temporal consistency in the final animation sequence.

\subsubsection{3D Head Rendering}
\label{head_rendering}
Using GAGAvatar~\cite{chu2024generalizable} as our 3D head modeling framework, we utilize the tracked FLAME parameters to render high-quality facial images that match our predefined pose sequence. We leverage this model to render the 3D head with precise control over pose and expression. The rendering process begins with the FLAME model, which generates 3D vertices based on the tracked shape, expression, pose, and eye parameters. We then employ a mesh renderer with a resolution of \( 512 \times 512 \) pixels, using the FLAME topology for face modeling where $\text{focal length}$ is set to $12.0$. This approach enables us to generate precisely controlled facial renderings that maintain the identity of the source image while adopting the pose and expression parameters from the target sequence.

\subsubsection{Face Fusion Process}
\label{face_fusion}
We selectively apply face fusion only to frames when the head rendering is front-facing. We determine this by analyzing eye landmark detection - specifically, when at least one eye is clearly visible and properly detected in the facial landmark set. This approach ensures face fusion is only applied to frames with reliable facial orientation, as the quality of renderings deteriorates for back-of-head views where no eyes are visible. After filtering, we perform structural similarity assessment~\cite{wang2017structural} and landmark-based warping~\cite{weisstein2004affine} with careful parameter tuning to ensure seamless integration.

First, we detect 68 facial landmarks using dlib~\cite{dlib} on both the diffusion-generated frame $I_{\text{orig}}$ and the rendered head image $I_{\text{head}}$ from GAGAvatar. Before applying the transformation, we validate the structural compatibility by computing a Procrustes disparity measure~\cite{goodall1991procrustes} between the landmark sets:
\begin{equation}
d(L_{\text{orig}}, L_{\text{head}}) = \sqrt{\frac{1}{n}\sum_{i=1}^{n} \|L_{\text{orig},i} - L_{\text{head},i}\|^2}
\end{equation}
where $L_{\text{orig}}$ and $L_{\text{head}}$ are the normalized landmark sets. We skip fusion when the disparity exceeds a threshold of $0.01$, preserving the original frame in cases where the structural alignment would produce unnatural results. For valid frames, we compute an affine transformation matrix through corresponding landmarks using:
\begin{equation}
M = \underset{M}{\arg\min} \sum_{i=1}^{68} \|M \cdot L_{\text{head},i} - L_{\text{orig},i}\|^2
\end{equation}
where $M$ is a $2 \times 3$ affine transformation matrix. This matrix is estimated using a partial affine model that preserves scale while allowing for rotation and translation, maintaining proportional facial features during transformation. The warped image is then computed by applying the transformation:
\begin{equation}
I_{\text{warp}} = T(I_{\text{head}}, M, (w, h))
\end{equation}
where $T$ represents the affine warping function that maps pixels from the source to destination image according to transformation $M$.

We then create a facial mask $\Omega$ by computing the convex hull~\cite{barber1996quickhull} of the landmarks to define the facial region:
\begin{equation}
\Omega = \text{convexHull}(L_{\text{orig}})
\end{equation}
Finally, we apply seamless cloning, a gradient-domain blending implementation of Poisson image editing~\cite{perez2023poisson}, centered at the face centroid $(c_x, c_y)$ with a blending factor $\alpha = 1.0$:
\begin{equation}
I_{\text{fused}} = \text{PoissonBlend}(I_{\text{warp}}, I_{\text{orig}}, \Omega, (c_x, c_y))
\end{equation}
This procedure solves the Poisson equation:
\begin{equation}
\min_{I} \int_{\Omega} \|\nabla I - \nabla I_{\text{warp}}\|^2 \, dx \, dy, \text{ subject to } I|_{\partial\Omega} = I_{\text{orig}}|_{\partial\Omega}
\end{equation}
The gradient-domain blending preserves boundary conditions from the original image while replacing interior gradients with those from the warped image. This approach maintains lighting conditions and color consistency across the boundary by solving for pixel values that create a smooth transition while matching gradient fields. The complete face fusion pipeline significantly reduces visible artifacts at the transition between the rendered face and the original image, allowing consistent identity preservation even under challenging viewpoints.

\subsection{Image Restoration Submodule}
To enhance the quality of video diffusion outputs, particularly in facial regions, we integrate a hybrid restoration pipeline based on BFRffusion~\cite{chen2024towards}. Our approach combines diffusion-based facial enhancement with background upsampling to improve overall visual fidelity while preserving identity-specific details.

The restoration workflow begins with face detection using RetinaFace~\cite{deng2020retinaface}, which accurately localizes facial regions in each frame. For aligned facial areas, we maintain a consistent face size of \( 512 \times 512 \) with a $1:1$ crop ratio. When processing non-aligned faces, we employ a landmark-based alignment process using a five-point facial landmark detector with an eye distance threshold of $5$ pixels to filter out low-quality detections.

Each detected face undergoes diffusion-based restoration using a latent diffusion model. The process follows a conditional diffusion sampling approach:
\begin{equation}
z_{t-1} = \frac{\sqrt{\alpha_{t-1}}z_t - \sqrt{1-\alpha_t}\epsilon_\theta(z_t)}{\sqrt{\alpha_t}} + \sqrt{1-\alpha_{t-1}}\epsilon_\theta(z_t)
\end{equation}
where $\alpha_t = \prod_{i=1}^t(1-\beta_i)$ and $\epsilon_\theta$ is the denoising network. We implement classifier-free guidance with a scale of $w = 3.5$:
\begin{equation}
\hat{\epsilon}_\theta(z_t) = (1+w)\epsilon_\theta(z_t) - w\epsilon_\theta(z_t, \emptyset)
\end{equation}
where $\epsilon_\theta(z_t, \emptyset)$ represents the unconditional prediction.

The diffusion sampling process uses $50$ DDIM steps with a latent shape of $\mathbb{R}^{4 \times 64 \times 64}$ for \( 512 \times 512 \) input images. The input facial image is first encoded to a latent representation through a VAE encoder, and the diffusion model progressively refines this representation before decoding it back to pixel space.

For background regions, we employ Real-ESRGAN~\cite{wang2021real} with an RRDBNet~\cite{gao2022efficient} architecture and a $2\times$ upsampling scale. The background upsampler processes images in tiles of \( 400 \times 400 \) pixels with $10$-pixel padding to handle high-resolution inputs efficiently while maintaining consistent quality across tile boundaries.

After separate processing of facial and background regions, we integrate the enhanced components using inverse affine transformations computed from the original facial alignment process. This creates a seamless composite where facial details are preserved and enhanced while maintaining natural transitions to background areas:
\begin{equation}
I_{\text{final}} = M_{\text{face}} \odot T^{-1}(I_{\text{face}}) + (1 - M_{\text{face}}) \odot I_{\text{bg}}
\end{equation}
where $T^{-1}$ represents the inverse transformation that maps the restored face back to its original position, and $M_{\text{face}}$ is the binary mask indicating facial regions.

This comprehensive image restoration approach significantly enhances the perceptual quality of generated frames, particularly improving fine facial details that may be lost or degraded during the initial video diffusion process. The integration of specialized facial and background processing ensures optimal quality across the entire frame while maintaining computational efficiency.

\subsection{Gaussian Avatar Submodule}
To transform our synthetic data into a high-quality, animatable 3D avatar, we employ a two-stage process: first, we fit an SMPL-X model to our synthetic data sequences, then we train a 3D Gaussian Splatting representation using the fitted parameters as guidance.

\subsubsection{SMPL-X Fitting Process}
Prior to training the 3DGS avatar, we employ a comprehensive fitting process to obtain accurate SMPL-X parameters from our synthetic data. This multi-stage process ensures that the avatar's geometry accurately reflects the subject's physical characteristics and articulation.
\vspace{0.2cm}

\noindent\textbf{Keypoint Extraction.} The fitting pipeline begins with pose and shape estimation. We utilize DWPose~\cite{yang2023effective} to extract 2D whole-body keypoints from each frame of our synthetic sequence. These keypoints provide critical information about body articulation across the sequence. The keypoints are represented as \( K \in \mathbb{R}^{J \times 3} \), where \( J = 133 \) includes $17$ body, $68$ face, and $42$ hand keypoints, with each keypoint having \((x, y, \text{confidence})\) values. We then employ MMPOSE~\cite{sengupta2020mm} with the RTMPose-L~\cite{jiang2023rtmpose} model for refinement, using a confidence threshold of $0.5$ to filter reliable detections.
\vspace{0.2cm}

\noindent\textbf{Initial Parameter Estimation.} For facial geometry, we leverage DECA~\cite{feng2021learning} to estimate initial FLAME parameters. The optimization uses perspective projection with focal length of $5000$ pixels and \( 1024 \times 1024 \) resolution textures. The FLAME parameters include shape coefficients \( \beta \in \mathbb{R}^{10} \), expression parameters \( \phi \in \mathbb{R}^{10} \), and pose parameters for jaw and eyes.

For body pose and shape, we incorporate Hand4Whole~\cite{moon2022accurate} with the configuration: focal length of $2000$, principal point at image center, and input shape of \( 256 \times 256 \). This process yields initial estimates for SMPL-X parameters: global orientation \(\theta_{\text{root}} \in \mathbb{R}^{3}\), body pose \(\theta_{\text{body}} \in \mathbb{R}^{21 \times 3}\), jaw pose \(\theta_{\text{jaw}} \in \mathbb{R}^{3}\), hand poses \(\theta_{\text{hands}} \in \mathbb{R}^{30 \times 3}\), and shape parameters \(\beta_{\text{shape}} \in \mathbb{R}^{10}\).
\vspace{0.2cm}

\noindent\textbf{Parameter Optimization.} These initial parameters are refined through an optimization process with multiple objectives. The primary loss function combines reprojection error, parameter regularization, and temporal smoothness:
\begin{equation}
L_{\text{fit}} = \lambda_{\text{kpt}} L_{\text{kpt}} + \lambda_{\text{reg}} L_{\text{reg}} + \lambda_{\text{temp}} L_{\text{temp}}
\end{equation}

The keypoint reprojection loss \(L_{\text{kpt}}\) measures the distance between projected model joints and detected 2D keypoints, weighted by detection confidence:
\begin{equation}
L_{\text{kpt}} = \sum_{i=1}^{J} c_i \| \Pi(J_i(\theta, \beta)) - K_i \|_2^2
\end{equation}
where \(\Pi\) is the perspective projection function, \(J_i(\theta, \beta)\) is the 3D position of joint \(i\), \(K_i\) is the corresponding 2D keypoint, and \(c_i\) is its confidence score.

The regularization term \(L_{\text{reg}}\) penalizes deviation from prior pose and shape distributions:
\begin{equation}
L_{\text{reg}} = \|\beta\|_2^2 + \sum_{j} \|\theta_j - \theta_{\text{mean}}\|_2^2
\end{equation}

The temporal consistency term \(L_{\text{temp}}\) enforces smooth transitions between frames:
\begin{equation}
L_{\text{temp}} = \sum_{t=1}^{T-1} \|\theta_t - \theta_{t+1}\|_2^2 + \|\beta_t - \beta_{t+1}\|_2^2
\end{equation}

The optimization uses the Adam optimizer~\cite{kingma2014adam} with learning rate $1 \times 10^{-3}$ and loss weights $\lambda_{\text{kpt}}=1.0$, $\lambda_{\text{reg}}=0.001$, and $\lambda_{\text{temp}}=0.1$. The optimization proceeds in two stages: first optimizing global position and orientation with 100 iterations, then refining all parameters with $200$ iterations.
\vspace{0.2cm}

\noindent\textbf{Parameter Smoothing.} To ensure temporal consistency and reduce jitter, we apply the same smoothing approach as used in our FLAME parameter tracking process in Section~\ref{flame_tracking}. Specifically, we employ Savitzky-Golay~\cite{john2021adaptive} filtering with a window length of $9$ frames and polynomial order of $2$. For rotation parameters, we utilize the identical quaternion-based smoothing procedure with continuity enforcement to handle sign flips.
\vspace{0.2cm}

\noindent\textbf{Segmentation and Depth Estimation.} We generate foreground masks using Segment Anything~\cite{kirillov2023segment} with the ViT-H backbone. The model uses keypoint-based prompting with valid keypoints as point coordinates, and a bounding box computed from these keypoints with an extension ratio of $1.2$. We also extract depth information using Depth Anything V2~\cite{yang2024depth} with the ViT-L backbone. The depth maps are normalized and aligned with the SMPL-X mesh using the following procedure:
\begin{equation}
\begin{gathered}
\text{scale} = \frac{\sigma(\text{depth}_{\text{pred},\text{fg}})}{\sigma(\text{depth}_{\text{smplx},\text{fg}})} \\
\text{depth}'_{\text{pred}} = \frac{\text{depth}_{\text{pred}}}{\text{scale}} \\
\text{depth}'_{\text{pred}} = \text{depth}'_{\text{pred}} - \mu(\text{depth}'_{\text{pred},\text{fg}}) + \mu(\text{depth}_{\text{smplx},\text{fg}})
\end{gathered}
\end{equation}
where \(\sigma\) and \(\mu\) represent standard deviation and mean of depth values, and \(\text{fg}\) indicates foreground regions.

The extracted SMPL-X parameters $\Phi = {\theta, \beta}$, together with corresponding image observations ${I_t}{t=1}^T$, foreground masks ${M_t}{t=1}^T$, and aligned depth maps ${D_t}_{t=1}^T$, constitute a multi-modal conditioning set that guides the optimization of our 3D Gaussian representation.

\subsubsection{3DGS Avatar Training Process}
With the fitted SMPL-X parameters and processed synthetic data, we proceed to train the 3DGS-based avatar~\cite{moon2024expressive}. The training begins by initializing the triplane representation~\cite{chan2022efficient} \( T \in \mathbb{R}^{32 \times 128 \times 128} \), encoding 3D features for both body and facial regions. Gaussian parameters, including positions \( \mathbf{V} \in \mathbb{R}^{N \times 3} \), colors \( \mathbf{C} \in \mathbb{R}^{N \times 3} \), and opacity \( \mathbf{O} \in \mathbb{R}^{N} \), are optimized through backpropagation with the following multi-objective loss function:
\begin{equation}
L = \lambda_{\text{RGB}} L_{\text{RGB}} + \lambda_{\text{SSIM}} L_{\text{SSIM}} + \lambda_{\text{LPIPS}} L_{\text{LPIPS}},
\end{equation}
where \( \lambda_{\text{RGB}} = 0.8 \), \( \lambda_{\text{SSIM}} = 0.2 \), and \( \lambda_{\text{LPIPS}} = 0.2 \) are the weights for the RGB reconstruction, structural similarity, and perceptual loss, respectively. The model is trained for 5 epochs with a batch size of 1, as required by the Gaussian splatting renderer.

The optimization process proceeds in two stages. During the warmup stage, Gaussian positions \( \mathbf{V} \) are updated using an adaptive learning rate:
\begin{equation}
\alpha_{\text{position}}(t) = \alpha_{\text{init}} \times \left( 1 - \frac{t}{T_{\text{max}}} \right) + \alpha_{\text{final}} \times \frac{t}{T_{\text{max}}},
\end{equation}
where \( \alpha_{\text{init}} = 1.6 \times 10^{-4} \), \( \alpha_{\text{final}} = 1.6 \times 10^{-6} \), and \( T_{\text{max}} = 30,000 \) iterations. Additional parameters, including opacity \( \mathbf{O} \), scale \( \mathbf{S} \), and feature parameters, are optimized with learning rates \( \alpha_{\text{opacity}} = 0.05 \), \( \alpha_{\text{scale}} = 0.005 \), and \( \alpha_{\text{feature}} = 0.0025 \), respectively.

Densification of the Gaussian distribution occurs between iteration $500$ and $15,000$, at intervals of $100$ iterations. Gaussians with opacity values below a threshold (\( \mathbf{O} < 0.005 \)) are pruned, and dense regions are refined using gradient-based adjustments. The pruning mechanism ensures efficient representation while preserving fidelity:
\begin{equation}
\mathbf{V}_{\text{new}} = \mathbf{V}_{\text{old}} - \eta \frac{\partial L}{\partial \mathbf{V}},
\end{equation}
where \( \eta \) is the learning rate and \( \frac{\partial L}{\partial \mathbf{V}} \) represents the gradient of the loss with respect to Gaussian positions.

A hierarchical learning approach progressively increases the spherical harmonic degree \( d_{\text{sh}} \) from $0$ to $3$ over the course of training. The training loop dynamically adjusts Gaussian parameters, leveraging an Adam optimizer with a learning rate of \( 1 \times 10^{-3} \) for the overall framework and parameter-specific rates for finer control. For our experiments, we employ the male SMPL-X~\cite{pavlakos2019expressive} model due to its superior performance in complex sequences. The entire pipeline runs on a single GPU, ensuring scalability and efficiency.

\begin{figure*}[ht]
\centering
\includegraphics[width=1.0\linewidth]{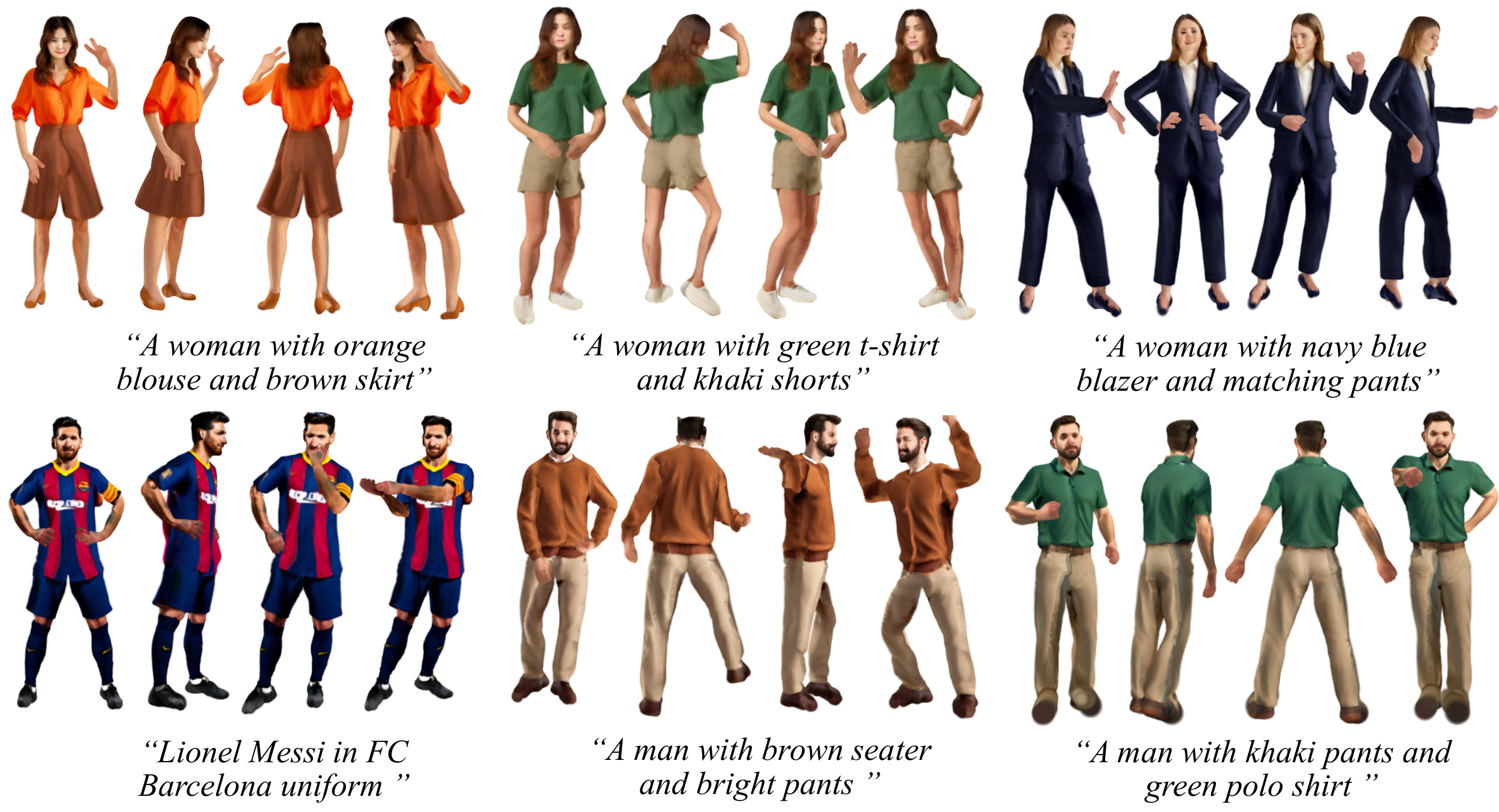}
\caption{\textbf{Text to 3D Avatar.} Our method enables the generation of animatable 3D avatars from text prompts. We show results for various textual descriptions processed through Flux-1 Dev~\cite{flux2024} for image generation, followed by our single-image to 3D avatar pipeline.}
\label{fig:app_text2avatar}
\end{figure*}

\section{Applications}
\label{supple-applications}
Our pipeline for 3D avatar generation from single images also serves as a basis for developing further creative applications. By integrating our core pipeline with contemporary text-to-image synthesis~\cite{flux2024} and image editing methods~\cite{liu2025step1x}, we expand its range of use. This section outlines two such applications: a text-to-3D avatar generation workflow in Section ~\ref{app_generation}, which enables the creation of animatable 3D characters from textual inputs, and a text-guided avatar editing application in Section ~\ref{app_editing}, which permits semantic modifications to an avatar's visual features using text prompts.

\subsection{Text to 3D Avatar}
\label{app_generation}
The generation of 3D avatars directly from textual descriptions is an area of interest, and various approaches ~\cite{liao2024tada, hong2022avatarclip, kolotouros2023dreamhuman, zhang2024rodinhd, jiang2023avatarcraft} have be explored. We first leverage the Flux-1 Dev~\cite{flux2024} model for text-to-image generation. This generated image then serve as input to our single-image to 3D avatar pipeline, producing  the 3D avatar that preserve the features described in the text, as shown in Figure~\ref{fig:app_text2avatar}. This integration extends the applicability of our framework to scenarios where photographic references are unavailable, thus broadening the scope of generative 3D human representation.

\subsection{Text-Guided 3D Avatar Editing}
\label{app_editing}
\begin{figure*}[ht]
\centering
\includegraphics[width=1.0\linewidth]{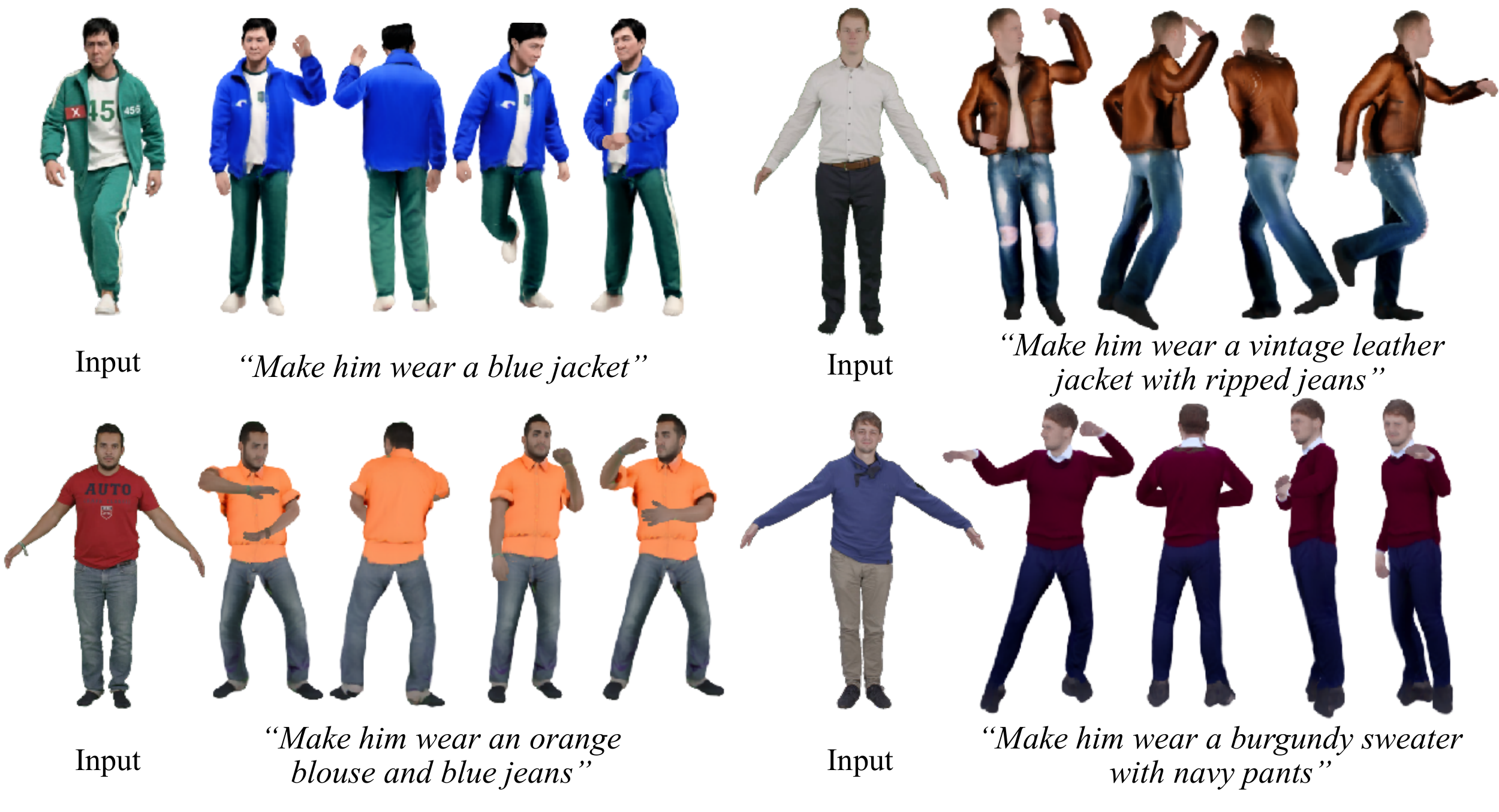} 
\caption{\textbf{Text-Guided 3D Avatar Editing.} Our framework enables semantic editing of 3D avatars using textual prompts. The resulting edited images are then processed by our single-image to 3D avatar pipeline, producing updated, animatable 3D avatars that reflect the specified textual edits.}
\label{fig:app_text_edit} 
\end{figure*}

Our pipeline enables text-guided 3D avatar editing~\cite{kim2024gala, zhang2024teca, liu2025gaussianavatar, cao2024dreamavatar} as illustrated in Figure~\ref{fig:app_text_edit}. By integrating Step1X-edit~\cite{liu2025step1x}, a text-based image editing diffusion model, we enable semantic modifications to the input image. Given an input image and a textual editing prompt, the diffusion model generates the modified reference image that incorporate the requested prompts. This edited image then proceeds through our standard single-image to 3D avatar pipeline, generating the modified 3D avatar that reflect the text-specified edits. This approach allows customizing avatar appearances without manual image manipulation or 3D modeling.

\vspace{-5pt}
\section{Failure Cases}
\label{supple-failure}
Despite the demonstrated effectiveness of SVAD, we observe several limitations that highlight opportunities for future research. Our analysis reveals three primary categories of failure cases that affect the quality and consistency of the generated avatars: issues stemming from segmentation artifacts in Section~\ref{failure:segmentation}, challenges in accurately modeling loose clothing deformation in Section~\ref{failure:clothing}, and inconsistencies observed in back view synthesis Section~\ref{failure:back_view}.

\subsection{Segmentation Artifacts}
\label{failure:segmentation} %
\begin{figure}[t]
\includegraphics[width=1.0\columnwidth, trim={2cm 0 0.2cm 1cm}]{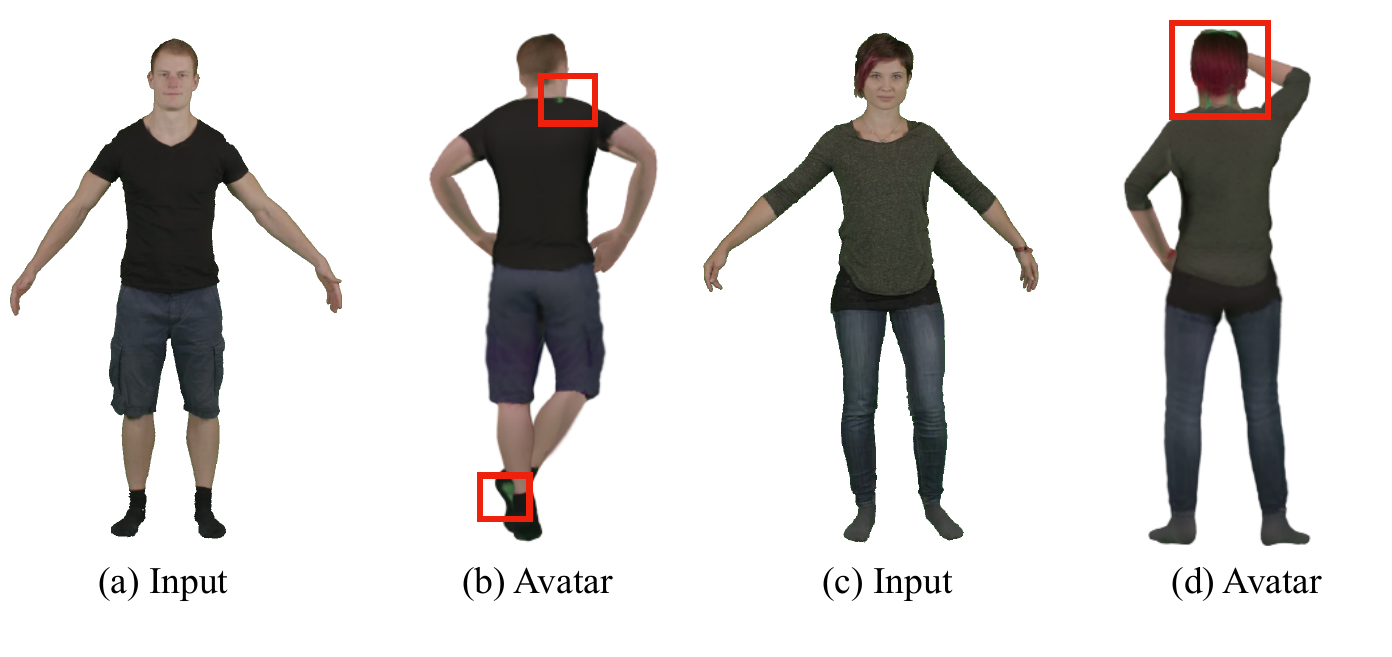}
\vspace{-25px}
\caption{\textbf{Failure case of segmentation artifacts.} When segmentation fails to properly separate the subject from the background, residual background elements become embedded in the avatar.}
\label{fig:limit_segmentation}
\vspace{3px}
\end{figure}

Limitations in the segmentation model~\cite{kirillov2023segment}, leveraged in preprocessing the generated synthetic data for 3D avatar training, can impact our pipeline's output, as illustrated in \cref{fig:limit_segmentation}. Background portions can be included in the training data for the 3D Gaussian representation. Such inaccuracies are most evident in posterior views of the reconstructed avatar, where artifacts from the original background are observed in the volumetric representation. These background elements remain visible during novel viewpoint synthesis, degrading the quality of rendered results. This shows our method's reliance on background segmentation, particularly for images with ambiguous foreground-background boundaries or similar color distributions.

\subsection{Loose Clothing Deformation}
\label{failure:clothing}
\begin{figure}[t]
\includegraphics[width=1.0\columnwidth, trim={2cm 0 0.2cm 1cm}]{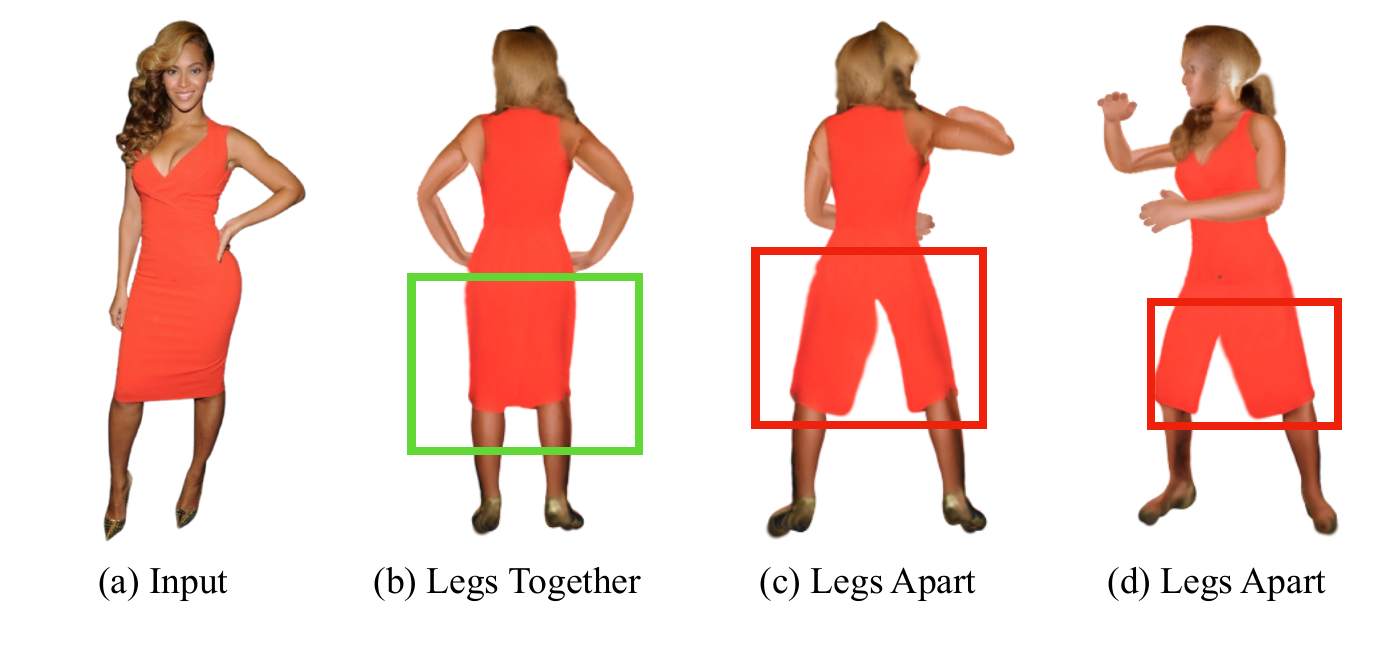}
\vspace{-25px}
\caption{\textbf{Failure case of loose clothing deformation. } Our method struggles with modeling loose garments like dresses due to limitations of the underlying parametric body model."}
\label{fig:limit_loose_clothing}
\vspace{3px}
\end{figure}

Modeling loose clothing, such as dresses and skirts, presents challenges for our method, as demonstrated in \cref{fig:limit_loose_clothing}. This problem arises from the parametric body model, which represents the human form as a close-fitting mesh and lacks explicit mechanisms for loose garments. While the rendered appearance can be plausible when the avatar's legs are proximate, fidelity degrades upon leg separation. The Gaussian splats, conditioned to align with to the parametric body surface, follow the leg geometry instead of preserving the garment's structure, leading to unrealistic deformations. This limitation indicates the need for modeling loose clothing deformation separately from the body mesh~\cite{PhysAavatar24}, especially for garments with flow-dependent behaviors that deviate from standard body topology.

\subsection{Back View Synthesis Inconsistency}
\label{failure:back_view}
\begin{figure}[t]
\includegraphics[width=1.0\columnwidth, trim={2cm 0 0.2cm 1cm}]{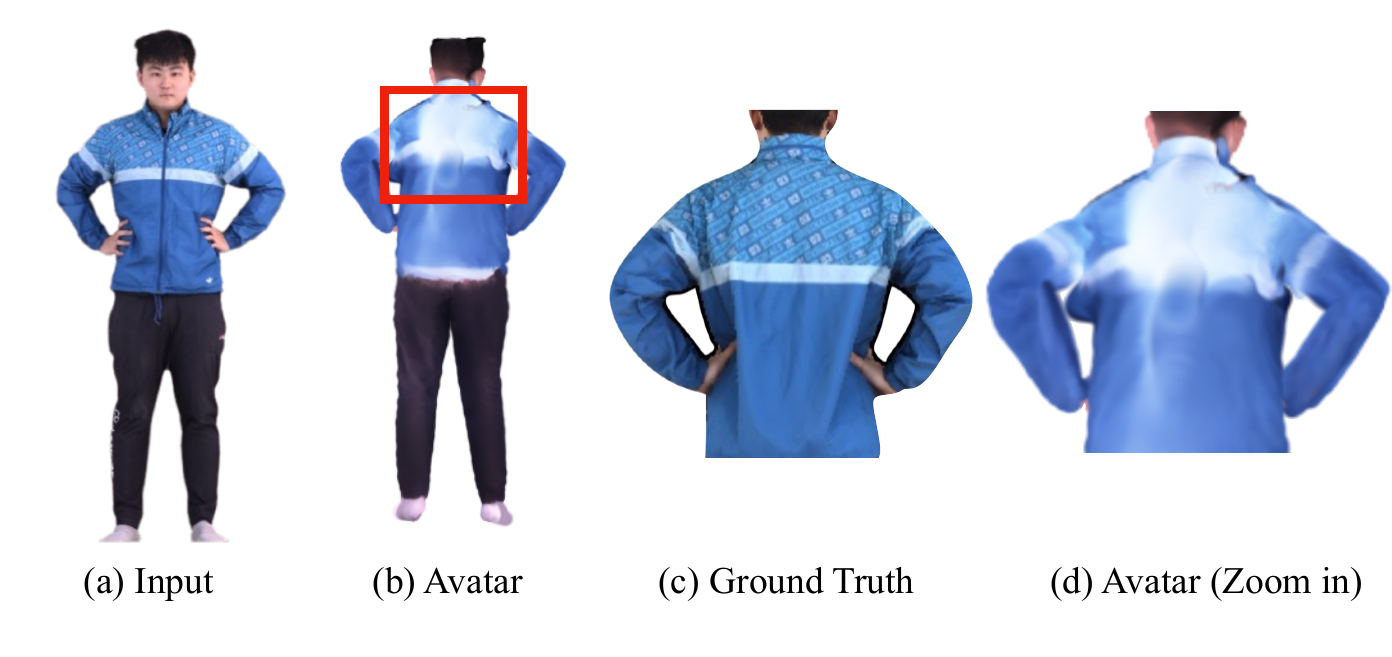}
\vspace{-25px}
\caption{\textbf{Failure case of back view synthesis.} The inherently ill-posed nature of single-image 3D generation results in degraded quality for unseen viewpoints."}
\label{fig:limit_backview}
\vspace{-5pt}
\end{figure}

A fundamental limitation is the ill-posed problem of single-image to 3D generation, which particularly affects back view synthesis in our approach. As shown in \cref{fig:limit_backview}, our method exhibits challenges in generating high-fidelity back views. While an input image provides strong frontal appearance cues, the synthesized back views can display reduced quality. Compared to ground truth renderings from similar viewpoints, our method often produce textures with lower fidelity, pattern inconsistencies, and blurred details. This behavior is attributed to two main factors: first, a potential bias in the video diffusion model towards frontal or near-frontal views during synthetic data generation, and second, the inherent ambiguity of inferring occluded geometry and appearance from a single perspective.

\section{Runtime Analysis}
\label{sec:time_analysis}
We evaluate the computational efficiency of our proposed pipeline. All experiments were conducted on a high-performance computing node equipped with an AMD EPYC 7742 64-Core Processor (128 logical cores) and an NVIDIA A100-SXM4-80GB GPU. Runtimes were measured using the average generation time of 30 randomly selected rendering samples from the THuman dataset~\cite{tao2021function4d}, with the single image resolution of $512\times512$ pixels. Table~\ref{tab:suppl_runtime} reports the average running time for each constituent component of our pipeline. It should be noted that runtime measurements may vary significantly across different hardware configurations; for example, performance on consumer-grade GPUs or older server architectures would likely result in longer processing times compared to our high-end experimental setup.

As detailed in Table~\ref{tab:suppl_runtime}, the total processing time for our pipeline is approximately 6 hours per subject. The 3DGS avatar training stage is identified as the primary computational bottleneck, accounting for approximately 3/4 of the total runtime. While this initial training phase represents a significant computational investment, the subsequently generated 3D avatars can be rendered in real-time. Future work will explore optimization techniques to reduce the computational demands, particularly for the most intensive stages of the pipeline.
\input{tables/suppl_runtime}

\section{Societal Impact}
\label{sec:societal_impact}
SVAD generates animatable 3D avatars from single input images, enhancing accessibility to personal digital representation. As with other generative AI technologies addressing human likeness, this capability presents both positive and negative societal implications. Positively, SVAD allows broader access to 3D avatar creation, enabling users to readily produce personalized digital representations for VR/AR, gaming, and enhanced online communication without specialized expertise. This can foster more engaging virtual interactions and broaden creative expression. Negatively, the ease with which an avatar can be generated from any single image, potentially without consent, poses considerable risks. Unauthorized 3D avatars could be exploited for deepfakes~\cite{westerlund2019emergence}, impersonation, harassment, or privacy violations. While comprehensive technical safeguards against all misuse are challenging, we will emphasize responsible application and the critical need for consent in any public dissemination of our work. We advocate for a multi-faceted approach to mitigate these risks including detection technologies, ethical guidelines, and legal frameworks and hope our research prompts further discussion on these vital societal considerations for the responsible development and deployment of such technologies.

\section{Acknowledgments.}
\label{sec:acknowledgments}
We thank Byungjun, Hyeonwoo, Taeksoo for valuable discussions and insights. We also extend our gratitude to SECERN AI for the generous compute grant for this work.

\begin{figure*}[ht]
\vspace{-60pt}
\centering
\includegraphics[width=0.94\linewidth]{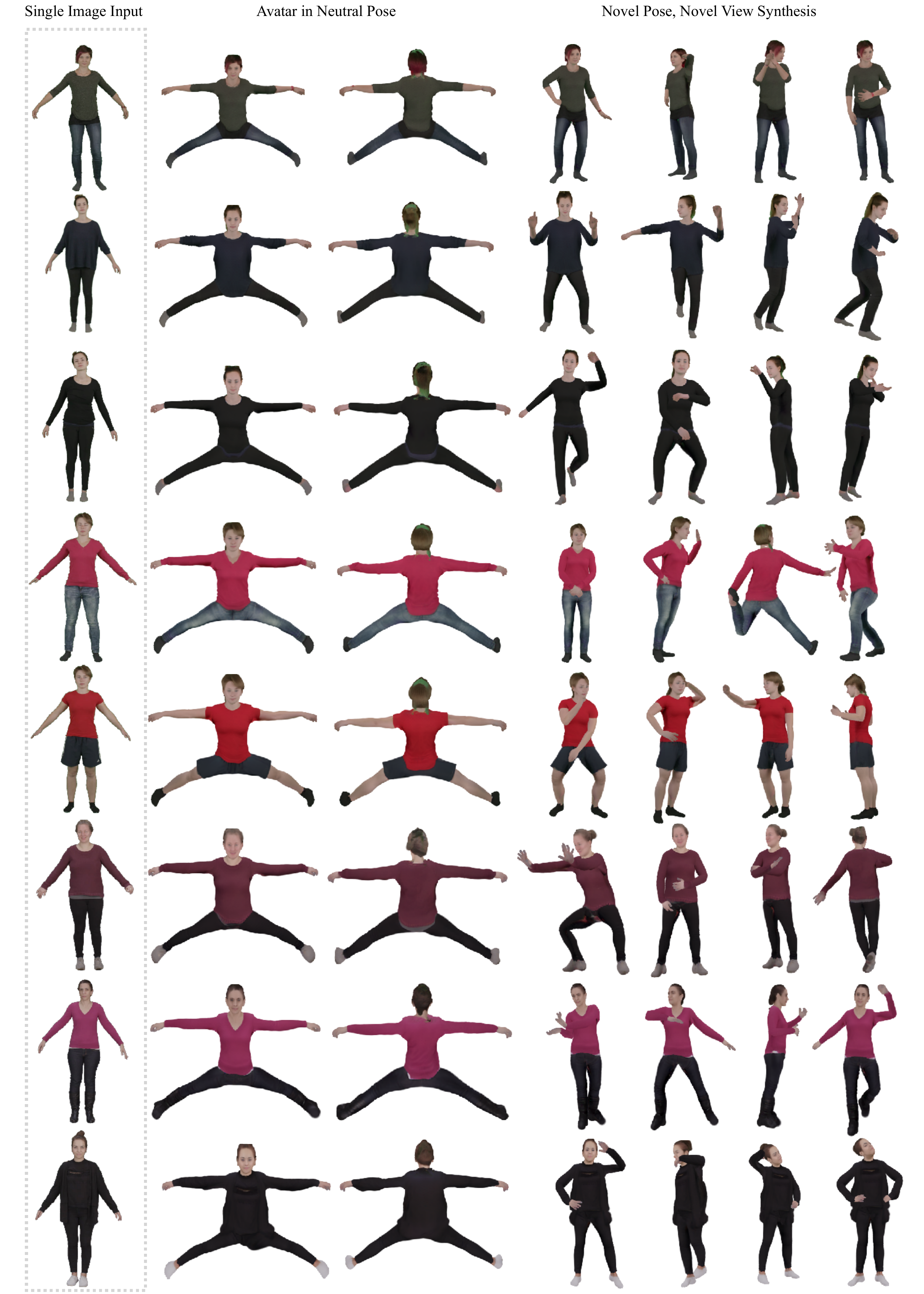}
\caption{\textbf{3D Avatars from People Snapshot~\cite{alldieck2018video} dataset} Our method successfully generates high-fidelity avatars for various subjects from a single input image, demonstrating robust identity preservation and consistent appearance across novel poses and viewpoints. \faSearchPlus\ ~\textbf{Zoom} in for more details. }
\label{fig:suppl_snapshot_female_avatars}
\end{figure*}

\begin{figure*}[ht]
\vspace{-60pt}
\centering
\includegraphics[width=0.95\linewidth]{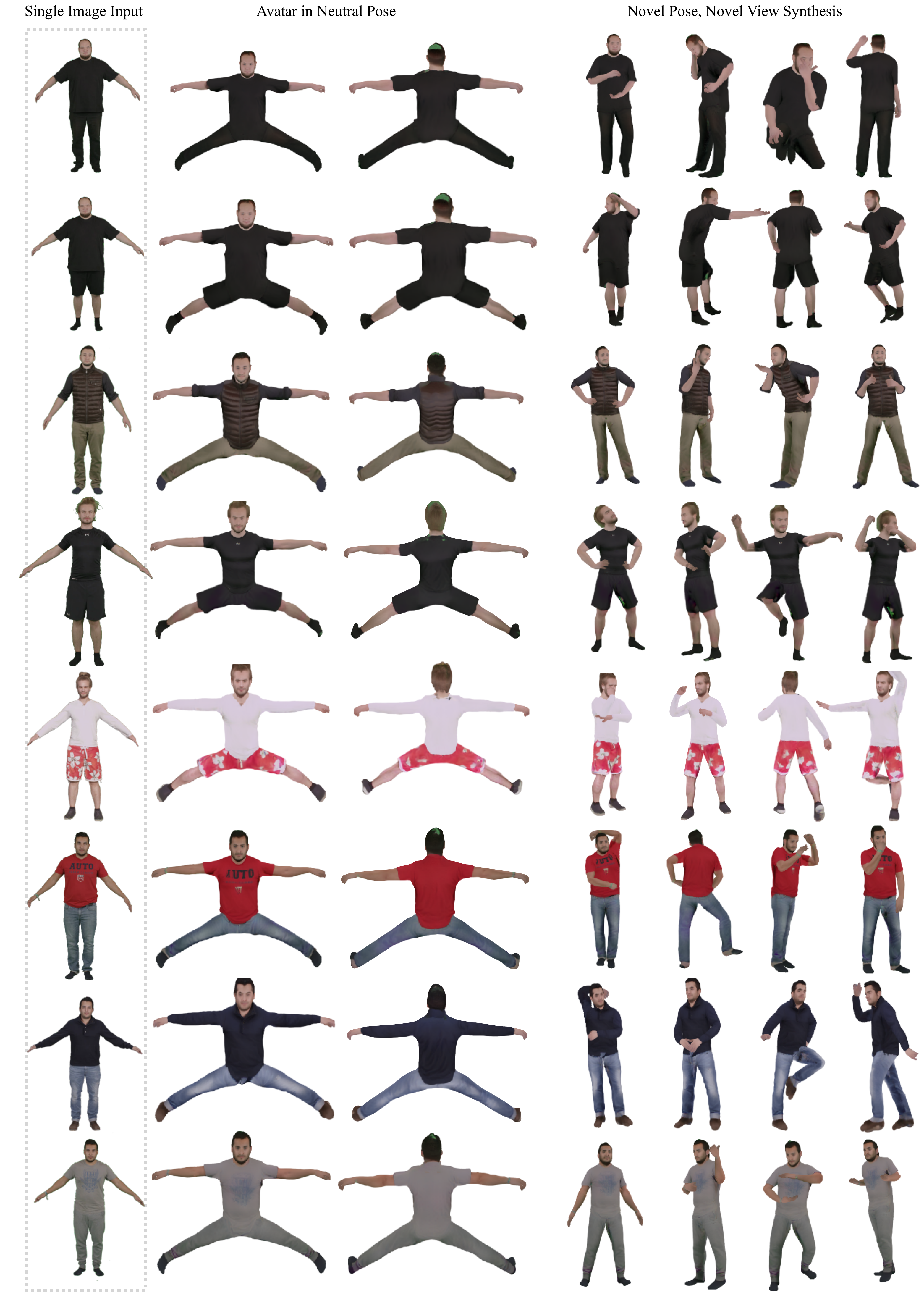}
\caption{\textbf{3D Avatars from People Snapshot~\cite{alldieck2018video} dataset.} SVAD enables creation of detailed and expressive avatars from a single image, accurately reproducing clothing details and facial features while maintaining realism in different poses. \faSearchPlus\ ~\textbf{Zoom} in for more details. }
\label{fig:suppl_snapshot_male_avatars}
\end{figure*}

\begin{figure*}[ht]
\vspace{-60pt}
\centering
\includegraphics[width=0.95\linewidth]{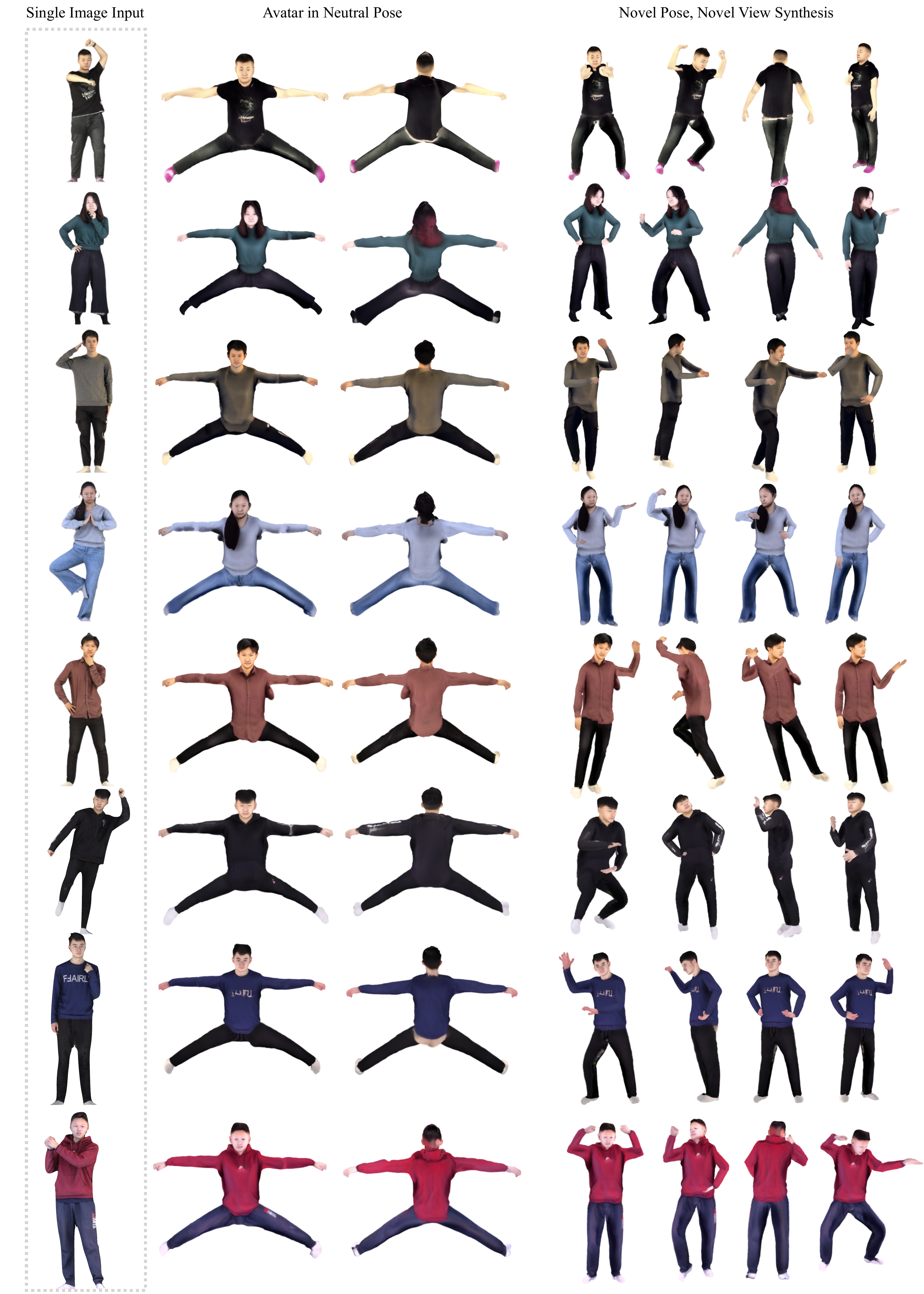}
\caption{\textbf{3D Avatars from the THuman~\cite{tao2021function4d} scan renderings.} Our approach generalizes well to the THuman dataset, producing realistic avatars with high geometric and texture fidelity. \faSearchPlus\ ~\textbf{Zoom} in for more details. }
\label{fig:suppl_thuman_avatars}
\end{figure*}

\begin{figure*}[ht]
\vspace{-65pt}
\centering
\includegraphics[width=0.95\linewidth]{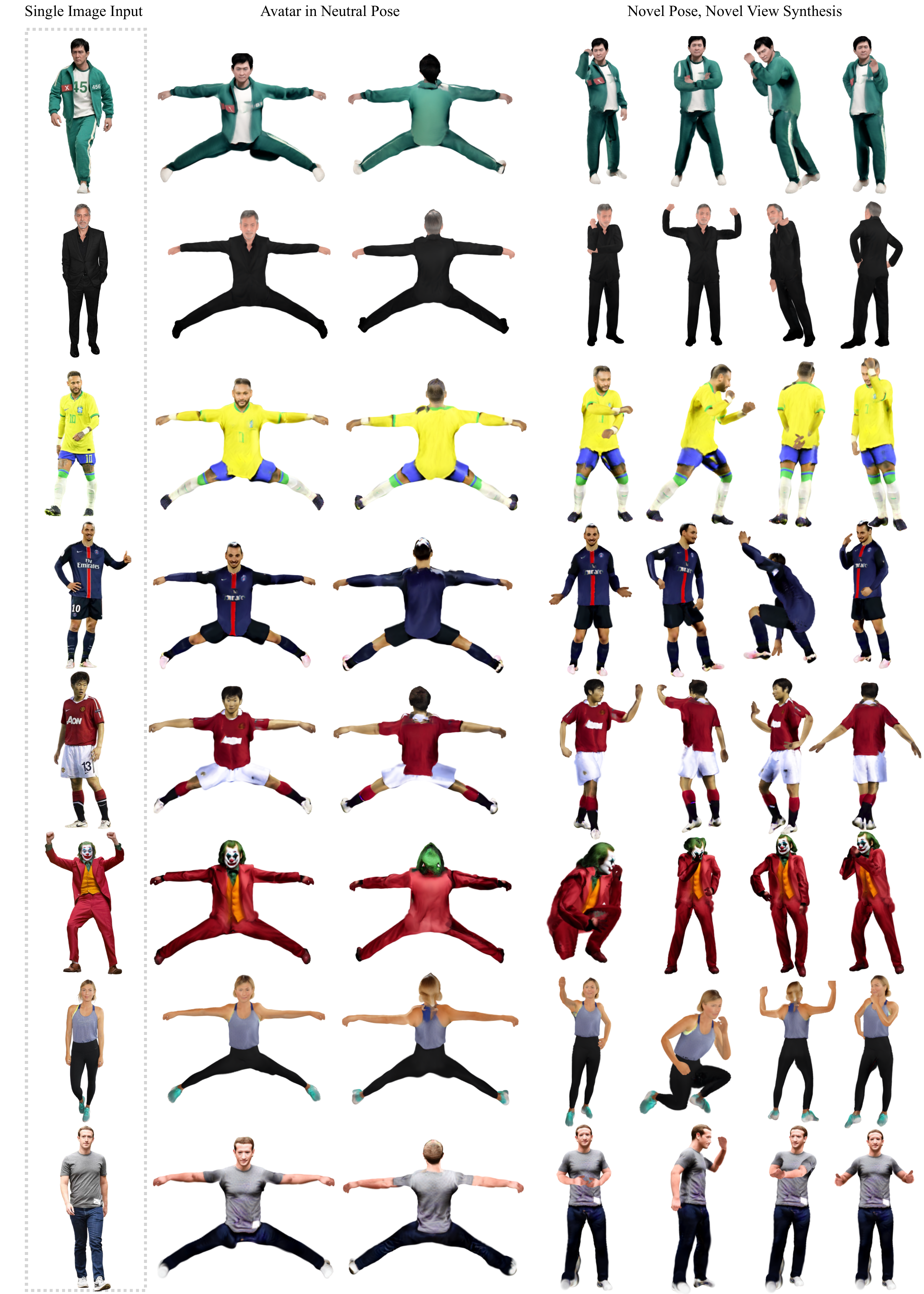}
\caption{\textbf{3D Avatars from Internet Images.} SVAD demonstrates strong generalization capability to in-the-wild images, successfully reconstructing recognizable 3D avatars of various celebrities from single unconstrained photographs. The method preserves distinctive appearance characteristics while enabling novel pose animation. \faSearchPlus\ ~\textbf{Zoom} in for more details. }
\label{fig:suppl_celebrity_avatars}
\end{figure*}

%% file: tables/suppl_runtime.tex
\begin{table}[t]
\centering
\small 
\setlength{\tabcolsep}{30pt} 
\begin{tabular}{@{}lr@{}} 
\toprule
\textbf{Pipeline Step} & \textbf{Time (min)}\\
\midrule
Video Diffusion Module &  18.33\\ 
Identity Preservation Module & 3.02 \\ 
Image Restoration Module &  11.00 \\
SMPL-X Fitting &  60.55 \\ 
3DGS Avatar Training &  273.23 \\ 
\midrule 
\textbf{Total} &  \textbf{366.13} \\ 
\bottomrule
\end{tabular}
\caption{\textbf{Running time analysis.} Average running time in minutes for each component of our pipeline.}
\label{tab:suppl_runtime}
\end{table}